%% file: paper.tex
\documentclass[sn-mathphys,Numbered]{sn-jnl}

\usepackage{afterpage}
\usepackage{bookmark}%
\usepackage{graphicx}%
\usepackage{multirow}%
\usepackage{amsmath,amssymb,amsfonts}%
\usepackage{amsthm}%
\usepackage{eucal}%
\usepackage{mathrsfs}%
\usepackage{mathtools}%
\usepackage[title]{appendix}%
\usepackage[dvipsnames]{xcolor}%
\usepackage{textcomp}%
\usepackage{manyfoot}%
\usepackage{booktabs}%
\usepackage{bm}%
\usepackage{algorithm}%
\usepackage{algorithmicx}%
\usepackage{algpseudocode}%
\usepackage{listings}%
\usepackage[inline]{enumitem}
\usepackage{overpic}
\usepackage{newfloat}
\usepackage{comment}
\usepackage{adjustbox}
\usepackage{layout}
\usepackage{mdframed}
\usepackage{framed}
\usepackage{siunitx}

\usepackage{tikz}
\usetikzlibrary{arrows.meta}
\usetikzlibrary{3d}
\usetikzlibrary{backgrounds}
\usetikzlibrary{calc}
\usetikzlibrary{positioning}
\usetikzlibrary{fit}
\usetikzlibrary{decorations.pathreplacing}
\usetikzlibrary{patterns}
\usetikzlibrary{shapes}

\usepackage{pifont}
\newcommand{\xxmark}{\textcolor{red}{\ding{53}}}%
\newcommand{\xmark}{\relax\ifmmode\mbox{\xxmark}\else\xxmark\fi}

\newcommand{\bnf}{\textsc{BayesNF}}

\usepackage{siunitx}
\usepackage[subrefformat=parens]{subcaption}
\DeclareCaptionLabelSeparator{sep2}{\hspace{2mm}}
\DeclareCaptionFormat{fmtTable}{\fontsize{8}{9.5}\selectfont#1#2#3\vphantom{y}}
\DeclareCaptionFormat{f8figure}{\fontsize{8}{9.5}\selectfont#1#2#3}
\DeclareCaptionLabelFormat{continued}{\bfseries#1~#2 (Continued)}
\DeclareFloatingEnvironment[name=Listing,listname={List of Listings}]{listing}
\usepackage[capitalize]{cleveref}
\crefname{line}{line}{lines}
\crefname{bbox}{Box}{Boxes}
\crefname{listing}{Listing}{Listings}

\makeatletter
\DeclareRobustCommand{\labelcrefrange}[2]{\@crefrangenostar{labelcref}{#1}{#2}}
\makeatother

\usepackage{nameref}

\newcommand{\s}{\mathbf{s}}
\newcommand{\br}{\mathbf{r}}
\DeclarePairedDelimiter\abs{\lvert}{\rvert}
\DeclarePairedDelimiter\set{\lbrace}{\rbrace}

\DeclarePairedDelimiter\floor{\lfloor}{\rfloor}
\DeclareMathOperator*{\argmax}{arg\,max}

\newcommand{\defas}{\coloneqq}

\newcommand{\simiid}{\overset{\rm iid}{\sim}}

\begin{document}

% \linenumbers

\title{\centering Scalable Spatiotemporal Prediction \\ with Bayesian Neural Fields}
\author*[1,2]{\fnm{Feras} \sur{Saad}}\email{fsaad@cmu.edu}
\author[2]{\fnm{Jacob} \sur{Burnim}}\email{jburnim@google.com}
\author[2]{\fnm{Colin} \sur{Carroll}}\email{colcarroll@google.com}
\author[2]{\fnm{Brian} \sur{Patton}}\email{bjp@google.com}
\author[2]{\fnm{Urs} \sur{Köster}}\email{ursk@google.com}
\author[2]{\fnm{Rif} \sur{A.~Saurous}}\email{rif@google.com}
\author[2]{\fnm{Matthew} \sur{Hoffman}}\email{mhoffman@google.com}

\affil[1]{\orgdiv{Computer Science Department}, \orgname{Carnegie Mellon University}, \orgaddress{\street{5000 Forbes Ave}, \city{Pittsburgh}, \state{PA} \postcode{15213}, \country{USA}}}
\affil[2]{\orgname{Google Research}, \orgaddress{\street{1600 Amphitheatre Pkwy}, \city{Mountain View}, \state{CA} \postcode{10587}, \country{USA}}}

% \begin{abstract}
\abstract{
Spatiotemporal datasets, which consist of spatially-referenced time series,
are ubiquitous in diverse applications, such as air pollution monitoring,
disease tracking, and cloud-demand forecasting.
As the scale of modern datasets increases, there is a growing need
for statistical methods that are flexible enough to capture complex
spatiotemporal dynamics and scalable enough to handle many observations.
This article introduces the Bayesian Neural Field (\bnf), a domain-general
statistical model that infers rich spatiotemporal probability distributions
for data-analysis tasks including forecasting, interpolation, and
variography.
\bnf{} integrates a deep neural network architecture for high-capacity
function estimation with hierarchical Bayesian inference for robust
predictive uncertainty quantification.
Evaluations against prominent baselines show that \bnf{} delivers
improvements on prediction problems from climate and public health data
containing tens to hundreds of thousands of measurements.
Accompanying the paper is an open-source software package
(\url{https://github.com/google/bayesnf}) that runs on GPU and
TPU accelerators through the \textsc{Jax} machine learning platform.
}
% \end{abstract}

\keywords{spatiotemporal prediction, Bayesian inference, deep neural networks, geostatistics}

\maketitle

\clearpage

\section*{Introduction}
\label{sec:introduction}

\emph{Spatiotemporal} data, which consists of measurements gathered at
different times and locations, is ubiquitous across diverse
disciplines.
Government bodies such as the European Environment
Agency~\citep{AirIndexEU} and United
States Environmental Protection
Agency~\cite{AirNowUS}, for example, routinely
monitor a variety of air quality indicators
($\mathrm{PM}_{10}$, $\mathrm{NO}_2$, $\mathrm{O}_3$, etc.)
in order to understand their ecological and public health impacts~\citep{Wang2006,Medina2004}.
As it is physically impossible to place sensors at all locations in a
large geographic area, environmental data scientists routinely develop
statistical models to predict these indicators at new locations or
times where no data is available~\citep{Huang2021,karagulian2019}.
%
% Accurately predicting these air quality indicators at new
% locations and time points is therefore a central problem faced by
% environmental data scientists.
%
Spatiotemporal data analysis also plays an important role in cloud
computing, where consumer demand for resources such as CPU, RAM, and
storage is driven by time-evolving macroeconomic factors and varies
across data center location.
Cloud service providers build sophisticated
demand-forecasting models to
determine prices~\citep{Niu2012}, perform load
balancing~\citep{Mishra2020}, save energy~\citep{Cao2012}, and
achieve service level agreements~\citep{Faniyi2015}.
Additional applications of spatiotemporal data analysis include
meteorology (forecasting rain volume~\citep{Sigrist2012} or wind speeds~\citep{Jung2014}),
epidemiology (``nowcasting'' active flu cases~\citep{Lu2019}),
and urban planning (predicting rider congestion patterns at metro stations~\citep{GanGan2020}).

Unlike traditional regression or classification methods in machine learning that operate
on independent and identically distributed (i.i.d.) data, accurate
models of spatiotemporal data must capture complex and highly nonstationary
dynamics in both the time and space domain.
For example, two locations twenty miles apart in California's central
valley may exhibit nearly identical temperature patterns, whereas two
locations only one mile apart in nearby San Francisco might have very
different microclimates; and these effects may differ depending on the
time of year.
Handling such variability across different scales is a key
challenge in designing accurate statistical models.
Another challenge is that spatiotemporal observations are typically driven by
unknown and noisily observed data-generating processes,
which requires models that report probabilistic predictions to account for
the aleortic and epistemic uncertainty in the data.

% \bmhead{Probabilistic Models of Spatiotemporal Data}
%
The dominant approach to spatiotemporal data modeling in statistics
rests on \textit{Gaussian processes}, a rich class of Bayesian nonparametric
priors on random functions~\citep{rasmussen2006,Cressie2011,Wikle2019}.
Consider a spatiotemporal field $Y(\s,t)$ indexed by
spatial locations $\s \in \mathbb{R}^d$ and time points $t\in\mathbb{R}$.
A typical Gaussian-process based ``prior probability'' distribution (used
in popular geostatistical software packages such as R-INLA~\citep{Rue2017} and
sdm-TMB~\citep{Anderson2022}) over the random field $Y$ is given by:
\begin{align}
\eta \sim \mathrm{GP}(0, k_\theta);\quad
F(\s,t) = h(x(\s,t); \beta) + \eta(\s,t);\quad
Y(\s,t) \sim \mathrm{Dist}(g(F(\s,t)), \gamma).
\label{eq:gp}
\end{align}
In \cref{eq:gp}, $\eta$ is a random function whose covariance over
space and time is determined by a \textit{kernel function}
$k_\theta((\s,t),(\s',t'))$ parameterized by $\theta$;
$x(\s,t)$ is a covariate vector
associated with index $(\s,t)$; $h$ is a mean function with parameters $\beta$
(e.g., for a linear function, $h(x; \beta) \coloneqq \beta'x$) of the latent field
$F$;
and $\mathrm{Dist}$ is a noise model (e.g., Normal, Poisson)
for the observations $Y(\s, t)$,
with index-specific parameter $g(F(\s,t))$
(where $g$ is a link function, e.g., $\exp$)
and global parameters $\gamma$.

Given an observed dataset
$\mathcal{D} \defas \set{Y(\s_1,t_1) = y_1, \dots, Y(\s_N,t_N) = y_N}$,
the inference problem is to determine the unknown parameters
($\theta$, $\beta$, $\gamma$), which in turn define a
\textit{posterior distribution} over the processes $(\eta, F, Y)$ given $\mathcal{D}$.
Advantages of the model~\labelcref{eq:gp} are \begin{enumerate*}[label=(\roman*)]

\item its flexibility, as $\eta$ is capable of representing highly complex
covariance structure; and

\item its ability to quantify uncertainty, as the posterior spreads its
probability mass over a range of functions and model parameters that are
consistent with the data.
\end{enumerate*}
Moreover, the model easily handles arbitrary patterns of missing data
by treating them as latent variables.
A number of recent articles have developed specialized Gaussian process
techniques for modeling rich spatiotemporal fields
\citep[e.g.,][]{Hamelijnck2021,Zhang2023,Anderson2022,Mangion2021,Banerjee2020}.

% \bmhead{Key Challenges}

Despite their flexibility, spatiotemporal models
based on Gaussian processes (such as \cref{eq:gp}) come with significant challenges.
The first is computational.
The simplest and most accurate posterior inference algorithms for
these models have a computational cost of $O(N^3)$, where $N$ is the
number of observations, which is unacceptably high in datasets with
tens or hundreds of thousands of observations.
Reducing this cost requires compromises, either on
the modeling side (e.g., imposing a discrete Markovian structure on the
model~\citep{Rue2017,Anderson2022}) or on the posterior-inference side (e.g.,
 approximating the true posterior with a simpler Gaussian process~\citep{Hamelijnck2021,Zhang2023,Banerjee2020}).
Either way, the resulting models have less expressive power and cannot
explain the data as accurately.
These approximations also involve delicate linear-algebraic
derivations or stochastic differential equations, which are
challenging to implement and apply to new settings.

The second challenge is expertise, where
the accuracy of model~\labelcref{eq:gp} on a given dataset is dictated
by key choices such as the covariance kernel $k_\theta$ and
mean function $h$.
Even for seasoned data scientists, designing these quantities is
difficult because it requires detailed knowledge about the
application domain.
Further, even small modifications to the model can impose large
changes to the learning algorithm, and so most software packages only support a small
set of predetermined covariance structures $k_\theta$ (e.g., separable Matérn
kernels, radial basis kernel, polynomial kernel) that are optimized
enough to work effectively on large datasets.

% \bmhead{This Work}

To alleviate these fundamental tensions, this article introduces the
\emph{Bayesian Neural Field} (\bnf)---a method that combines the
scalability of deep neural networks with many of the attractive
properties of Gaussian processes.
\bnf{} is built on a Bayesian neural network model
\citep{Neal1996} that maps from multivariate space-time coordinates to
a real-valued field.
The parameters of the network are assigned a prior distribution, and
as in Gaussian processes, conditioning on observed data induces a
posterior over those parameters (and in turn over the entire field).
Because inference is performed in ``weight space'' rather than ``function
space'', the cost of analyzing a dataset grows linearly with the
number of observations, as opposed to cubically for a Gaussian process.
Because \bnf{} is a hierarchical model (\cref{fig:network}),
it naturally handles missing data as latent variables
and quantifies uncertainty over parameters and predictions.
And because \bnf{} defines a field over continuous space--time,
it can model non-uniformly sampled data, interpolate in space,
and extrapolate in time to make predictions at novel coordinates.

Our description of \bnf{} as a neural ``field'' is inspired by the
recent literature on neural radiance fields
\citep[NeRFs;][]{mildenhall2021nerf,hoffman2023probnerf} in computer vision.
A key discovery that enabled the success of NeRFs is that neural
networks are biased towards learning functions whose Fourier spectra
are dominated by low frequencies, and that this bias can be corrected
by concatenating sinusoidal positional encodings to the raw spatial
inputs \citep{tancik2020fourier}.
To ensure that our \bnf{} model assigns high prior probability to
data that includes both low- and high-frequency variation, we append
Fourier features to the raw time and position data that are fed to the
network.
In \textit{\nameref{sec:methods}}, we show that these
Fourier features, coupled with learned scale factors and convex combinations
of activation functions, improve \bnf{} models' ability
to learn flexible and well-calibrated distributions of spatiotemporal data.
Incorporating sinusoidal seasonality features lets \bnf{} models
make predictions based on (multiple) seasonal effects as well.
Taken together, these characteristics enable state-of-the-art
performance in terms of point predictions and 95\% prediction intervals on
diverse large-scale spatiotemporal datasets, without the need to
heavily customize the \bnf{} model structures on a per-dataset basis.

% \bmhead{Related Work}
\bnf{} belongs to a family of emerging techniques that
leverage deep neural networks with hierarchical Bayesian models
for spatiotemporal data analysis---a thorough survey of these advances is given in \citet{Wikle2023}.
Our method is inspired by limitations of existing deep neural network
approaches for probabilistic prediction in spatiotemporal data.
For example, the Bayesian spatiotemporal recurrent neural networks
introduced in \citet{McDermott2019} require the data to be observed at
a fixed spatial grid and regular discrete-time intervals.
In contrast, \bnf{} is defined over continuous space-time
coordinates, enabling prediction at novel locations and in datasets
with irregularly sampled time points.
The deep ``Empirical Orthogonal Function'' model~\citep{Amato2020} is
a powerful exploratory analysis tool but is less useful for
prediction: it cannot handle missing data, make predictions at new
time points, or deliver uncertainty estimates.
Additional methods in this category include Bayesian neural networks
that are highly task oriented---e.g., for analyzing power
flow~\citep{Gao2024}, wind speed~\citep{Liu2020}, or floater intrusion
risk~\citep{Wang2022}.
These methods leverage domain-specific architectures designed
specifically for the analysis problem at hand, and do not aim to
provide software libraries that are easy for practitioners to apply in
new spatiotemporal datasets beyond the application domain.
In contrast, a central goal of \bnf{} is to provide a domain-general
modeling tool that is easily applicable to the same type of datasets
as the Gaussian process model~\labelcref{eq:gp}, without the need to
redesign substantial parts of the probabilistic model or network
architecture for each new task.

% \item deep neural network warpings of inputs to Gaussian
% processes~\citep{Vu2023} include hard-to-tune
% decisions about how to order the observations and how to
% define their nearest-neighbor sets.
%
Neural processes~\citep{Garnelo2019} also integrate deep neural networks
with probabilistic modeling, but are based on a graphical model
structure that is fundamentally difficult to apply to spatiotemporal
datasets.
In particular, because neural processes aim to ``meta-learn'' a prior
distribution over random functions, the authors note it is essential
to have access to a large number of independent and identically
distributed (i.i.d.) datasets during training.
%
% From the NP paper: To learn such a distribution over random
% functions, rather than a single function, it is essential to train
% the system using multiple datasets concurrently, with each dataset
% being a sequence of inputs x1:n and outputs y1:n
%
However, most spatiotemporal data analyses are based on only a single
real-world dataset (e.g., those in \cref{table:datasets}) where there is no
notion of sharing statistical strength across multiple
i.i.d.~observations of the entire field.

Graph neural networks (GNNs), surveyed in~\citet{Ming2023}, are another
popular deep-learning approach for spatiotemporal prediction which
have been particularly useful in settings such as analyzing traffic
or population-migration patterns.
These models require as input a graph describing the connectivity
structure of the spatial locations, which makes them less appropriate
for spatial data that lack such discrete connectivity structure.
Moreover, the requirement that the graph be fixed makes it harder for
GNNs to interpolate or extrapolate to locations that are not included
in the graph at training time.
The \bnf{} model, on the other hand, operates over continuous space,
and is therefore more appropriate for spatial data without known discrete
connectivity structure.
In addition, as noted in \citet{Ming2023}, GNNs have not yet
been demonstrated on probabilistic prediction tasks, and we are
unaware of the existence of open-source software libraries based on
GNNs that can easily handle the sparse datasets in
\cref{table:datasets}.

\section*{Results}
\label{sec:results}

\subsection*{Model Description}
\label{sec:results-model}

\input{fig-network-vertical}

Consider a dataset
$\mathcal{D} = \set{y(\s_i,t_i) \mid i=1,\dots,N}$
of $N$ spatiotemporal observations, where
$\s_i \in \mathcal{S} \subset \mathbb{R}^d$
denotes a $d$-dimensional spatial coordinate and
$t_i \in \mathcal{T} \subset \mathbb{R}$ denotes a time index.
For example, if the field is observed at
longitude-latitude coordinates in discrete time,
then $\mathcal{S} = (-180, 180] \times [-90, 90] \subset \mathbb{R}^2$
and $\mathcal{T} = \set{1, 2, \dots,}$.
If the field also incorporates an altitude dimension, then
$\mathcal{S} \subset \mathbb{R}^3$.
We model this dataset as a realization
$\set{Y(\s_i,t_i)=y(s_i,t_i), 1 \le i \le n}$ of a
random field $Y: \mathcal{S} \times \mathcal{T} \to \mathbb{R}$ over
the entire spatiotemporal domain.
Following the notation in \citet{Wikle2023},
we describe the field using a hierarchical Bayesian model:
\begin{align}
&&\mbox{Observation Model:}\; &[Y(\cdot) \mid F(\cdot), \Theta_y], \label{eq:model-obs}\\
&&\mbox{Process Model:}\; &[F(\cdot) \mid x(\cdot), \Theta_f], \label{eq:model-process} \\
&&\mbox{Parameter Models:}\; &[\Theta_y, \Theta_f]. \label{eq:model-params}
\end{align}
In this notation, upper case letters denote random quantities, Greek
letters denote model parameters, lower case letters denoted non-random
(fixed) quantities, and square brackets $[\cdot]$ denote
(yet-to-specified) probability distributions.
The distribution of the observable random variables $Y(\s,t)$ is
parameterized by global parameters $\Theta_y$ and an
unobservable (latent) spatiotemporal field $F(\s,t)$.
In turn, $F(\s,t)$ is parameterized by a set of random global
parameters $\Theta_f$ and a collection
$x(\s,t) = [x_1(\s,t), \dots, x_m(\s,t)]$ of $m$ fixed covariates
associated with index $(\s,t)$.

\Cref{listing:bnf} completes the definition of \bnf{}
by showing specific probability distributions
for the model~\labelcrefrange{eq:model-obs}{eq:model-params}.
\Cref{fig:network} shows a probabilistic-graphical-model representation of
a \bnf{} model with $H=3$ layers, which takes a spatiotemporal index $(\s,t)$ at the
input layer and generates a realization $Y(\s,t)$ of the observable field
at the output layer.
At a high level, the input layer transforms the spatiotemporal coordinates
$(\s,t)$ into a fixed set of spatiotemporal covariates, which include
linear terms, interaction terms, and Fourier features in time and
space.
The second layer performs a linear scaling of these covariates using a
learnable scale factor---this layer aims to avoid the need for the
practitioner to manually specify how to appropriately scale the data,
which is known to heavily influence the learning
dynamics~\citep{LeCun2012}.
Next, the hidden layers of the network contain the usual dense
connections, except that the activations are specified as a learnable
convex combination of ``primitive'' activations, such as rectified
linear units (relu), exponential linear unit (elu), or hyperbolic
tangent (tanh).
The goal of these convex combinations is to automate the discovery of
the covariance structure in the field, given that activation functions
correspond directly to covariance of random functions defined by
Bayesian neural networks~\citep{pearce2020}.
At the final layer, the output of the feedforward network is used to
parameterize a probability distribution over the observable field
values, which serves to capture the fundamental aleortic uncertainty
in the noisy data.
Epistemic uncertainty in \bnf{} is expressed by assigning prior
probability distributions to all learnable parameters, such as
covariate scale factors; connection weights, biases, and their
variances; and additional parameters of the observation distribution.

We next describe the components of this process in sequence from
inputs to outputs in more detail.
This description defines a \emph{prior}
distribution over Bayesian Neural Fields---in \nameref{sec:methods}
we discuss ways of inferring the \textit{posterior}
over the random variables defined in \cref{listing:bnf}.

\bmhead{Spatiotemporal Covariates}

Letting $(\s,t) = ((s_1,\dots,s_d), t)$ denote a generic index in the field,
the covariates $[x_1(\s,t), \dots, x_m(\s,t)]$ may include
the following functions:
\begin{align}
&\set{t, s_1, \dots, s_d} && \mbox{\slshape Linear Terms}
  \label{eq:cov-linear}
  \\
&\set{ts_1, \dots, ts_d} && \mbox{\slshape Temporal-Spatial Interactions}
  \label{eq:cov-interaction-temporal}
  \\
&\set{s_is_j ; 1 \le i < j \le d} && \mbox{\slshape Spatial-Spatial Interactions}
  \label{eq:cov-interaction-spatial}
  \\
&\set{(\cos(2\pi h/p t), \sin(2\pi h/p t));
  p \in \mathcal{P}, h\in \mathcal{H}^{\rm t}_p} && \mbox{\slshape Temporal Seasonal Features}
  \label{eq:cov-seasonal}
  \\
&\set{(\cos(2\pi 2^h s_i), \sin(2\pi 2^h s_i));
  1 \le i \le d, h \in \mathcal{H}^{\rm s}_i} && \mbox{\slshape Spatial Fourier Features}
  \label{eq:cov-fourier}
\end{align}

\input{fig-generative}

The linear and interaction covariates~\labelcrefrange{eq:cov-linear}{eq:cov-interaction-spatial}
are the usual first and second-order effects used in
spatiotemporal trend-surface analysis models~\citep[Sec.~3.2]{Wikle2019}.
In \cref{eq:cov-seasonal}, the temporal seasonal features are defined by a set
$\mathcal{P} = \set{p_1, \dots, p_{\ell}}$ of $\ell$ seasonal periods, where each
$p_i$ has harmonics $\mathcal{H}^{\rm t}_{p_i} \subset \set{1,2,\dots, \floor{p_i/2}}$
for $i=1,\dots,\ell$.
For example, if the time unit is hourly data and there are $m=2$
seasonal effects (daily and monthly), the corresponding
periods are $p_1 = 24$ and $p_2=730.5$, respectively.
Non-integer periodicities handle seasonal effects that have varying
duration in the time measurement unit (e.g, days per month or weeks per
year).
The \nameref{sec:methods} section discusses how to
construct appropriate seasonal features for a variety of time
units and seasonal effect combinations.
In \cref{eq:cov-fourier}, the spatial Fourier features for coordinate
$s_i$ are determined by a set $\mathcal{H}^{\rm s}_i \subset \mathbb{N}$ of
additional frequencies that capture periodic structure in the $i$-th
dimension $(i=1,\dots,d)$.
These covariates correct for the tendency of neural networks to learn
low-frequency signals \citep{tancik2020fourier}: the empirical evaluation
in the next section confirms that their
presence greatly improves the quality of learned models.
Covariates may also include static (e.g., ``continent'') or dynamic (e.g.,
``temperature'') exogenous features, provided they are known at all
locations and time points in the training and testing datasets.

\bmhead{Covariate Scaling Layer}
Scaling inputs improves neural network
learning~\citep[e.g.,][]{LeCun2012}, but determining the appropriate strategy
(e.g., z-score, min/max, tanh, batch-norm, layer-norm, etc.) is challenging.
\bnf{} uses a prior distribution over scale factors to learn these
quantities as part of Bayesian inference within the overall
probabilistic model.
In particular, the next stage in the network is a width-$m$ hidden layer
$h^0_i(\s,t) = e^{\xi_i^0} x_i(\s,t)$ obtained by randomly scaling each
of the $m$ covariates $x(\s,t)$, where $e^{\xi^0_i}$
is a log-normally distributed scale factor (for $i=1,\dots,m$).

\bmhead{Hidden Layers}
The model contains $L + 1 \ge 1$ hidden layers, where layer $l$
has $N^{\ell}$ units $h^{\ell} = (h^{\ell}_1,\dots,h^{\ell}_{N^{\ell}})'$
(for $l=1,\dots,L$).
These hidden units are derived from $N^{\ell}$
\textit{pre-activation} units $z^\ell = 1/\sqrt{N^{\ell-1}}\Omega^\ell {h^{\ell-1}} + \beta^\ell$
where $\Omega^\ell = [\omega^\ell_{ij}; 1 \le i \le N^{\ell}, 1 \le j \le N^{\ell-1}]$
is a random $N^{\ell} \times N^{\ell-1}$ weight matrix and
$\beta^\ell = (\beta^\ell_1, \dots, \beta^\ell_{N^{\ell}})'$ a random bias term.
The network parameters $\omega^\ell_{ij}$ and $\beta^\ell_i$ are drawn
i.i.d.\ $N(0, \sigma^\ell)$, where the variance $\sigma^\ell = \ln(1+e^{\xi^\ell})$
is a learnable parameter whose prior is obtained by applying
a softplus transformation to $\xi^\ell \sim N(0,1)$.
The $1/\sqrt{N^{\ell-1}}$ term ensures the network has a well-defined
Gaussian process limit as the number of hidden units $N^{\ell} \to \infty$~\citep{Neal1996}.

In addition to the covariate scaling layer,
\bnf{} departs from a traditional Bayesian neural network by using
$A^\ell \ge 1$ activation functions $(u^\ell_1, \dots, u^\ell_{A^\ell})$ at hidden
layer $l$, instead of the usual $A^\ell = 1$.
For example, the architecture shown in \cref{fig:network} uses $A^\ell = 2$
where $u^\ell_1$ is the hyperbolic tangent (tanh) and $u^\ell_2$
is the exponential linear unit (elu) activation (where $l=1,2$).
Each \textit{post-activation} unit $h^\ell_i$ (for $i=1,\dots,N^{\ell})$
is then a random convex combination of the activations
$u^\ell_1(z^\ell_i), \dots, u^\ell_{A^\ell}(z^\ell_i)$,
where the coefficient of $u^\ell_j$ is the
output of a softmax function ${e^{\gamma^{\ell}_j}}/{\sum_{k=1}^{N_d} e^{\gamma^{\ell}_k}}$
whose $j$-th input is $\gamma^\ell_j \sim N(0,1)$
(for $j=1,\dots,A^\ell$).
The activation function governs the overall covariance properties of
the random function defined by a Bayesian neural
network~\citep{Neal1996,pearce2020}.
By specifying the overall activation at each layer as a learnable
convex combination of $A^\ell$ ``basic'' activation functions (e.g.,
tanh, relu, elu), \bnf{} aims to automate the process of selecting an
appropriate activation and in turn the covariance structure within the
random field.

Finally, the latent stochastic process $F(\s,t)$ is defined as the
pre-activation unit $z^{L+1}_1$ of layer $L+1$, which has exactly $N^{L+1}=1$ unit.
We let $\Theta_f$ denote all $n_f$ random network parameters
in \cref{listing:bnf} and denote the prior as $\pi_f$.
Further, the notation $F_{\theta_f}(\s,t)$ denotes the
(deterministic)  value of the process $F$ at index $(\s,t)$
when $\Theta_f = \theta_f$.

\bmhead{Observation Layer}
The final layer connects the stochastic
process $F(\s,t)$ with the observable spatiotemporal field
$Y(\s,t)\sim \mathrm{Dist}(F(\s,t); \Theta_y)$
through a noise model that captures
aleatoric uncertainty in the data.
The parameter vector $\Theta_y = (\Theta_{y,1}, \dots, \Theta_{y,n_y})$ is
$n_y$-dimensional and has a prior $\pi_y$.
There are many choices for this distribution, depending on the field
$Y(\s,t)$; for example,
\begin{align}
Y(\s,t) &\sim \mathrm{Normal}(F(\s,t), \Theta_{y,1}),
\label{eq:dist-normal}\\
Y(\s,t) &\sim \mathrm{StudentT}_{\Theta_{y,2}}(F(\s,t), \Theta_{y,1}),
\label{eq:dist-studentT} \\
Y(\s,t) &\sim \mathrm{Poisson}(e^{F(\s,t)}), \label{eq:dist-poisson}
\end{align}
which correspond to a Gaussian noise model with mean $F(\s,t)$ and variance
$\Theta_{y,1}$ ($n_y=1$), a
StudentT model with location $F(\s,t)$, scale $\Theta_{y,1}$
and $\Theta_{y,2}$ degrees of freedom ($n_y=2$);
and a Poisson counts model with rate $\exp{F(\s,t)}$ ($n_y=0$), respectively.
A key design choice in these observation distributions is that certain
parameters such as $\Theta_{y,1}$ in \cref{eq:dist-normal} or $\Theta_{y,1}, \Theta_{y,2}$
in \cref{eq:dist-studentT} are not index-specific but rather shared across
all inputs, which serves to mitigate the model's sensitivity to over-fitting
noise fluctuations from high-frequency Fourier features.

\bmhead{Posterior Inference and Querying}

Let $P(\Theta_f, \Theta_y, Y)$ be the joint probability
distribution over the parameters and observable field in
\cref{listing:bnf}. The posterior distribution given $\mathcal{D}$ is
\begin{align}
\begin{aligned}[t]
&P(\theta_f, \theta_y \mid \set{Y(\s_i,t_i)=y(\s_i,t_i)}_{i=1}^N)\\
&\qquad\propto
  \left(\prod_{i=1}^{n_f} \pi_f(\theta_{f,i})\right)
  \left(\prod_{i=1}^{n_y} \pi_y(\theta_{y,i})\right)
  \prod_{i=1}^n \mathrm{Dist}(y(\s_i,t_i); F_{\theta_f}(\s_i,t_i), \theta_y)
\label{eq:posterior-distribution}
\end{aligned}
\end{align}
While the right-hand side of \cref{eq:posterior-distribution} is tractable
to compute, the left-hand side cannot be normalized or sampled from exactly.
In the \nameref{sec:methods-posterior} section of \nameref{sec:methods}, we
discuss two approximate posterior inference algorithms for \bnf{}:
maximum a-posteriori ensembles and variational inference ensembles.
They each produce a collection of parameters
$\set{(\theta_f^i, \theta_y^i)}_{i=1}^M \approx P(\Theta_f, \Theta_y \mid \mathcal{D})$
drawn from an approximation to the posterior~\labelcref{eq:posterior-distribution}.
The \nameref{sec:methods-queries} subsection of \nameref{sec:methods}
discusses how these posterior samples be used
to compute \textit{point predictions}
$\hat{y}(\s_*,t_*)$ of the spatiotemporal field at a novel index $(\s_*, t_*)$
and the associated \textit{prediction intervals}
$[\hat{y}_{\rm low}(\s_*,t_*), \hat{y}_{\rm hi}(\s_*,t_*)]$
for a given level $\alpha \in (0,1)$ (e.g., $\alpha=95\%$).

\subsection*{Prediction Accuracy on Scientific Datasets}
\label{sec:results-evaluation-pred}

\subparagraph{Datasets}

To quantitatively assess the effectiveness of \bnf{} on challenging
prediction problems, we curated a benchmark set comprised of six
publicly available, large-scale spatiotemporal datasets that together
cover a range of complex empirical processes:

\begin{enumerate}[itemsep=4pt, topsep=4pt,leftmargin=*]
\item Daily wind speed (\unit{\km\per\hour}) from the Irish Meteorological Service~\citep{Haslett1989}.\\
1961-01-01 to 1978-12-31; 12 locations; 78,888 observations, 0\% missing.

\item Daily particulate matter 10 (PM10, \unit{\ug\per\meter^3}) air quality in Germany
from the European Environment Information and Observation Network~\citep{Pebesma2012}.\\
1998-01-01 to 2009-12-31; 70 locations; 149,151 observations, 52\% missing.

\item Hourly particulate matter 10 (PM10, \unit{\ug\per\meter^3}) from the London Air Quality Network~\citep{Hamelijnck2021}.\\
2018-12-31 to 2019-03-31; 72 locations; 144,570 observations, 7\% missing.

\item Weekly chickenpox counts (thousands) from the Hungarian National
Epidemiology Center~\citep{Chickenpox2021}\\
2005-01-03 to 2014-12-29; 20 locations; 10,440 observation, 0\% missing.

\item Monthly accumulated precipitation (\unit{\mm}) in Colorado and surrounding areas
from the University Corporation for Atmospheric Research~\citep{Precipitation2010}.\\
1950-01-01 to 1997-12-01; 358 locations; 134,800 observations, 35\% missing.

\item Monthly sea surface temperature (\unit{\degreeCelsius}) anomalies in the Pacific Ocean
from the National Oceanic and Atmospheric Administration Climate Prediction Center~\citep{Wikle2019}\\
1970-01-01 to 2003-03-01; 2261 locations; 902,139 observations, 0\% missing.
\end{enumerate}

\input{fig-datasets}

\Cref{table:datasets} summarizes key statistics of these datasets.
\Cref{fig:datasets} shows snapshots of the observed data at a fixed
point in time (\cref{fig:datasets-space}) and in space
(\cref{fig:datasets-time}), highlighting the complex statistical
patterns (e.g., nonstationarity and periodicity) in the underlying
fields along these two dimensions.
Five train/test splits were created for each benchmark.
Each test set contains $(\# \mathrm{locations}) / (\# \mathrm{splits})$
locations, holding out the 10\% most recent observations.

\bmhead{Baselines}
The prediction accuracy on the benchmark datasets in
\cref{table:datasets} using \bnf{} is compared to several
state-of-the-art baselines.
This evaluation focuses specifically on baseline methods that
\begin{enumerate*}[label=(\roman*)]
\item have high-quality and widely used open-source implementations;
\item can generate both point and interval predictions; and
\item are directly applicable to new spatiotemporal datasets (e.g.,
those in \cref{table:datasets}) without the need to redevelop
substantial parts of the model.
\end{enumerate*}
The baselines are:

\begin{enumerate}[itemsep=4pt, topsep=4pt,leftmargin=*]
\item \textsc{StSVGP}: Spatiotemporal Sparse Variational Gaussian Process~\citep{Hamelijnck2021}.
  This method handles large datasets (i.e., linear time scaling in the
  number of time points) by leveraging a state-space
  representation based on stochastic partial differential equations and Bayesian
  parallel filtering and smoothing on GPUs.
  Parameter estimation is performed using natural gradient variational inference.

\item \textsc{StGBoost}: Spatiotemporal Gradient Boosting Trees~\citep{Pedregosa2011}.
  Prediction intervals are estimated by minimizing the quantile loss using
  an ensemble of 1000 tree estimators.
  As this baseline is not a typical time series model, the same covariates
  $[x_1(\s,t), \dots, x_m(\s,t)]$~\labelcrefrange{eq:cov-linear}{eq:cov-fourier} provided
  to \bnf{} are also provides as regression inputs.

\item \textsc{StGLMM}: Spatiotemporal Generalized Linear Mixed Effects Models~\citep{Anderson2022}.
  These methods handle large datasets by integrating
  latent Gaussian-Markov random fields with stochastic partial differential equations.
  Parameter estimation is performed using maximum marginal likelihood inference.
  Three observation noise processes are considered:
  \subitem IID: Independent and identically distributed Gaussian errors.
  \subitem AR1: Order 1 auto-regressive Gaussian errors.
  \subitem RW: Gaussian random walk errors.

\item \textsc{NBEATS}: Neural Basis Expansion Analysis~\citep{Oreshkin2020}.
  This baseline employs a ``window-based'' deep learning
  auto-regressive model where future data is predicted over a
  fixed-size horizon conditioned on a window of previous observations
  and exogenous features. The model is configured with indicators for
  all applicable seasonal components---e.g., hour of
  day, day of week, day of month, week of year, month---as well as
  trend and seasonal Fourier features. The method contains a large
  number of numeric hyperparameters which are automatically tuned
  using the NeuralForecast~\cite{NeuralForecast2024} package.
  Prediction intervals are estimated by minimizing quantile loss.

\item \textsc{TSReg}: Trend Surface Regression with Ordinary Least Squares (OLS)~\citep[Sec.~3.2]{Wikle2019}.
  The observation noise model is Gaussian with maximum likelihood
  estimation of the variance. As with \textsc{StGBoost}, the
  regression covariates are identical to those provided to \bnf{}.

\item \bnf: Bayesian Neural Field; using variational and maximum a-posteriori inference.
% \item \bnfns: Same as above, without spatial Fourier features~\labelcref{eq:cov-fourier}.
\end{enumerate}
We also attempted to use the fixed-rank kriging (\textsc{Frk})
method~\citep{Mangion2021}, but were unable to
perform inference over noise parameters for spatiotemporal data.
Taken together, the baselines provide broad coverage over recent
statistical, machine learning, and deep learning methods for
large-scale prediction.
All methods were run on a TPU v3-8 accelerator, which consists of 8
cores each with 16 GiB of memory.
Additional evaluation details are described in \nameref{sec:methods}.

\bmhead{Quantitative Results}

\Cref{table:evaluation} shows accuracy and runtime results for all
baselines and benchmarks.
Point predictions are evaluated using root-mean square error
(RMSE~\labelcref{eq:rmse}) and mean absolute error
(MAE~\labelcref{eq:mae}) and 95\% prediction intervals are evaluated
using the mean interval score (MIS~\labelcref{eq:mis}), averaged over
all train/test splits.
The final column shows the wall-clock runtime in seconds that each
method was run.
While runtime cannot be perfectly aligned due to variety of learning
algorithms used and their iterative nature, the wall-clock numbers
show that all baselines were run for sufficiently long to ensure a
fair comparison.
\Cref{fig:predictions} compares predictions on held-out data at one
representative spatial location in each of the six benchmarks.
We discuss several takeaways from these results.

\bnf{} using VI is the strongest baseline in 12/18 cases
followed by \bnf{} using MAP: it is tied with VI in 3/18
cases (Precipitation) and superior in 3/18 cases
(Sea Surface Temperature).
In 2/18 cases (Chickenpox; MAE and RMSE) errors from the \bnf{}
methods are slightly higher than the \textsc{StGLMM}
(AR1) baseline, although the running time of the latter is $\sim$4x higher.
The most apparent improvements of \bnf{} occur in the Wind Speed,
Precipitation, and Sea Surface Temperature datasets, shown
qualitatively in rows 1, 5, 6 of \cref{fig:predictions}.
Results using additional ablations are discussed
in the \nameref{sec:methods-ablations} subsection of \nameref{sec:methods}.
Combined with \cref{table:evaluation}, these results highlight
the expressive modeling capacity of \bnf{} models, their ability to
accurately quantify predictive uncertainty, and the benefit of using
spatial embeddings to capture high-frequency signals in the data.

\afterpage{
\input{fig-evaluation}
\clearpage
\input{fig-predictions}
\clearpage
}

While predictions from \textsc{StSVGP} generally follow the overall ``shape'' of the
held-out data, the mean and interval predictions are not
well calibrated (\cref{fig:predictions}, second column).
\textsc{StSVGP} requires several modeling trade-offs to ensure linear-time
scaling in the number of time points, including the use of
Mat{\'e}rn kernels (which cannot express effects such as seasonality) and
kernels that are separable in time and space.
Additional difficulties include manually selecting the
number of spatial inducing points and complex algorithms needed to
optimize their locations.
\textsc{StSVGP} runs out of memory on the Sea Surface Temperature benchmark
(1 million observations).

The \textsc{StGLMM} methods (AR1, IID, RW) fail to complete on 4/6 benchmarks.
The scaling characteristics are also unpredictable: for example,
\textsc{StGLMM} runs on Air Quality 2 (144,570 observations) but fails on
Wind Speed (78,888 observations).
On the two datasets they can handle (rows 3 and 4 of
\cref{fig:predictions}), the \textsc{StGLMM} methods are highly competitive
on Chickenpox and not competitive on Air Quality 2, with the AR1 error model
delivering the lowest errors.

\textsc{StGBoost} delivers
reasonable prediction intervals but its point predictions underfit
(\cref{fig:predictions}, third column).
It has a high computational cost because \begin{enumerate*}[label=(\roman*)]

\item a large number of estimators is needed to obtain accurate predictions
(using 1000 estimators provided statistically significant improvements over
500 estimators in 17/18 benchmarks);

\item three models must be separately trained from scratch:
one model to predict the mean and two models to predict upper and lower
quantiles.
\end{enumerate*}
Whereas \bnf{} uses a single learned distribution for all queries,
\textsc{StGBoost} trains different models for different queries,
which does not guarantee probabilistically coherent answers.

\textsc{NBEATS} is only competitive on the Sea Surface Temperature
benchmark, where it is the next-best baseline after \bnf{}.
Its runtime on this benchmark is 3x--4x faster than \bnf{} due to
automatic early stopping.
The method fails to deliver predictions on the Precipitation benchmark
because the training and test datasets contain time series that
are too sparse to handle; e.g., the number of observed timepoints is
smaller than the auto-regressive window size or prediction horizon.
The prediction errors on the remaining three benchmarks are high
even though all the seasonal effects were added to the model,
suggesting that either \begin{enumerate*}[label=(\roman*)]
  \item the model is not able to effectively leverage
  spatial correlations for cross time-series learning; or
  \item the hyperparameter tuning algorithm does not converge to
  sensible values within the allotted time.
\end{enumerate*}

\textsc{TSReg} requires less than 1 second to train, but does not
capture any meaningful structure and produces poor predictions.
Using LASSO or ridge regression instead of OLS did not improve the results.
\textsc{TSReg} uses identical covariates to \bnf{} but performs much
worse, highlighting the need to capture nonlinear dependencies in the
data for generating accurate forecasts.

\subsection*{Analyzing German Air Quality Data}
\label{sec:results-evaluation-air}

Atmospheric particulate matter (PM10) is an key indicator of air
quality used by governments worldwide, as these particles can induce
adverse health effects when inhaled into the lungs.
Accurate predictions of PM10 values at novel points in space and
time within a geographic region can help decision makers characterize
pollution patterns and inform public health decisions.

We explore predictions from \bnf{} on the German Air
Quality dataset~\citep{Pebesma2012}, which contains daily PM10
measurements from 70 stations between 1998-01-01 and 2009-12-31.
We infer a \bnf{} model for this dataset with depth $H=2$; weekly,
monthly, and yearly seasonal effects~\labelcref{eq:cov-seasonal}; and harmonics
$\mathcal{H}^{\rm s}_1 = \mathcal{H}^{\rm s}_2 = \set{1,\dots,4}$ for the spatial
Fourier features~\labelcref{eq:cov-fourier}.
The distribution of $Y$ given the stochastic process $F$ is a StudentT~\labelcref{eq:dist-studentT}
truncated to $\mathbb{R}_{\ge 0}$

\afterpage{
\input{fig-germany}
\clearpage
\input{fig-variogram}
\clearpage
}

\bmhead{Spatial and Temporal Interpolation}
\Cref{fig:air-data} shows the PM10 observations at
2003-02-01, 2005-01-01, 2005-04-01, and 2007-01-01, where roughly 50\%
of the stations do not have an observed measurement at a given point in time.
\Cref{fig:air-spatial} shows the median PM10 predictions
$y_{0.5}(\s_*,t_*)$~\labelcref{eq:quantile} interpolated at a grid of 10,000
novel spatial indexes $(\s_*,t_*)$ within Germany.
\Cref{fig:air-spatial-95} shows the width
$\hat{y}_{\rm hi}(\s_*,t_*) - \hat{y}_{\rm low}(\s_*,t_*)$
of the inferred 95\% prediction interval.
These plots reflect the spatiotemporal structure captured by \bnf{}
and identify coordinates within the field with low and high predictive
uncertainty about air pollution.
The axis-aligned artifacts in \cref{fig:air-spatial},
where predictions are consistent along certain thin regions, are a
result of the spatial Fourier features \labelcref{eq:cov-fourier}. How well
these artifacts reflect the true behavior can be empirically
investigated by obtaining PM10 measurements at the novel locations along
these regions.
\Cref{fig:air-temporal} shows the observed and median predicted PM10 values
across all time points at four stations with the highest missing data
rates:
DEBWO31, southwest Germany, 51\% missing;
DEBB056, northeast Germany, 84\% missing;
DEBU034, northwest Germany, 99\% missing;
DESL008, west Germany, 89\% missing.
PM10 trajectories predicted by \bnf{} at time points where data is missing
reproduce the temporal patterns at
time points with observed data, which include high frequency periodic
variation and irregular, spatially correlated jumps.

\bmhead{Variography}

The accuracy of PM10 predictions in \cref{fig:air-temporal} cannot be
quantitatively assessed because the ground-truth values are not known
at the predicted time points.
However, we can gain more insight into how well the learned spatiotemporal
field matches the observed field by comparing the empirical and inferred
semi-variograms.
The semi-variogram $\gamma$ of a process $Y$ characterizes the joint
spatiotemporal dependence structure; it is defined as
\begin{align}
2\gamma(\mathbf{h},\tau) = \mathrm{Var}\left[ Y(\s+\mathbf{h},t+\tau) - Y(\s,t) \right]
  && (\mathbf{h} \in \mathcal{S}, \tau \in \mathcal{T}),
  \label{eq:semivariogram}
\end{align}
where the choice of $\s \in \mathcal{S}, t \in \mathcal{T}$ is
arbitrary (e.g., $(\s, t) = (\mathbf{0}, 0)$, under the assumption
that only the displacements in time and space affect the dependence
\citep[Sec.~2.4.2]{Wikle2019}).

The surface plots in \cref{fig:variogram} compare the empirical
semi-variogram (left) computed at the 70 observed stations with the
inferred semi-variogram (right) computed at 70 uniformly chosen random
locations within Germany,
for distances $\mathbf{h}\in [0, 1000]$ kilometers and time lags
$\tau \in \set{0,\dots,10}$ days.
The agreement between these two plots suggests that \bnf{}
accurately generalizes the spatiotemporal dependence structure from
the observed locations to novel locations in the field.
The lower two panels in \cref{fig:variogram} show the empirical (solid
line) and inferred (dashed line) semi-variograms, separately for each
of the 10 time lags $\tau$.
The difference between the semi-variograms is highest for
$\tau \in \set{0,1,2}$ days, suggesting that the learned model
is expressing relatively smooth phenomena and assuming that
the high-frequency day-to-day variance is due to unpredictable
independent noise.
The differences between the semi-variograms become small for $\tau > 2$ days,
which suggests that \bnf{} effectively
captures these longer-term temporal dependencies.

\section*{Discussion}
\label{sec:discussion}

This article proposes a probabilistic approach to scalable
spatiotemporal prediction called the Bayesian Neural Field.
The model combines a deep neural network architecture for high-capacity
function approximation with hierarchical Bayesian modeling for accurate
uncertainty estimation over complex spatiotemporal fields.
Posterior inference is conducted using stochastic ensembles of
maximum a-posteriori estimation or variationally trained surrogates,
which are easy to apply and deliver well-calibrated 95\% prediction
intervals over test data.
The results in \cref{fig:runtime} confirm that
quantifying uncertainty using MAP or VI ensembles is superior to
performing maximum-likelihood estimation (MLE), which ignores the
parameter priors.
While these inference methods are approximate in nature and are not
guaranteed to match the true posterior, the \bnf{} model is a deep neural
network where interpreting parameters such as weights and biases
is not of inherent interest to a practitioner in a given data analysis
task.
Rather, we expect \bnf{} to be most useful in cases where the predictive
calibration is more relevant.
Additional advantages of \bnf{} are its relative simplicity,
ability to handle missing data, and ability to learn a full
probability distribution over arbitrary space-time indexes
within the spatiotemporal field.

Evaluations against prominent statistical and machine learning baselines on
large-scale datasets show that \bnf{} delivers significant improvements in
both point and interval forecasts.
The results also show that combining periodic effects in the temporal
domain with Fourier features in the spatial domain enables \bnf{} to
capture spatiotemporal patterns with multiple (non-integer)
periodicity and high-frequency components.
As a domain-general method, \bnf{} can produce strong results on
multiple datasets without the need to hand-design the model from scratch
each time or apply dataset-specific inference approximations.
For a representative air quality dataset,
the semi-variograms inferred by \bnf{} evaluated at novel spatial
locations agree with the empirical semi-variogram computed at observed
locations, which highlights the model's ability to generalize
well in space and time.

Practitioners across a spectrum of disciplines---from meteorology to urban
studies and environmental informatics---are in need of more scalable
and easy-to-use statistical methods for spatiotemporal prediction.
A freely available implementation of \bnf{} built on the
\textsc{Jax} machine learning platform, along with user documentation and
tutorials, is available at \url{https://github.com/google/bayesnf}.
We hope these materials help practitioners obtain strong
\bnf{} models for many spatiotemporal problems that
existing software cannot easily handle.

The approach discussed in this paper opens several avenues to future work.
While Bayesian Neural Fields are designed to minimize the user's
involvement in constructing a predictive model, further improvements
can be achieved by enabling domain experts to incorporate specific
statistical covariance structure that they know to be present.
It is also worthwhile to explore applications of \bnf{} for modeling the
residuals of causal or mechanistic laws in physical systems where
there exist strong domain theories of the average data-generating
process, but poor models of the empirical noise process.
Another promising extension is using \bnf{} models to handle not only
``geostatistical'' datasets, in which the measurements are
point-referenced in space, but also ``areal'' or ``lattice'' datasets,
where the measurements represent aggregated quantities over a
geographical region.
While areal dataests are often converted to geostatistical datasets by
using the centroid of the region as the representative point, a more
principled approach would be to compute the integral of a
Bayesian Neural Field over the region.
Finally, \bnf{} can be generalized to handle multivariate
spatiotemporal data, where each spatial location is associated with
multiple time series that contain within-location and across-location
covariance structure.
Effectively handling such datasets will even further broaden the scope
of problems that \bnf{} can solve.

\section*{Methods}
\label{sec:methods}

\subsection*{Posterior Inference}
\label{sec:methods-posterior}

Let $P(\Theta_f, \Theta_y, Y)$ denote the joint probability
distribution over the parameters and observable field in
\cref{listing:bnf}.
The posterior distribution is given by
\cref{eq:posterior-distribution} in the main text.
We describe two approximate posterior inference algorithms for \bnf{}.
In these sections, we define
$\Theta = (\Theta_f, \Theta_y)$, $\theta = (\theta_f,\theta_y)$
and $\br = (\s,t)$.

\paragraph{Stochastic MAP Ensembles}
A simple approach to uncertainty quantification is based on the
``maximum a-posteriori'' estimate:
\begin{align}
% \theta^* &= \argmax_{\theta}\set*{\log P(\theta_f \theta_y \mid \set{Y(\br_i)=y(\br_i)}_{i=1}^N)}
\theta^* &= {\textstyle\argmax_\theta}\set*{\log P(\theta_f, \theta_y, \set{Y(\br_i)=y(\br_i)}_{i=1}^N)}.
% \nonumber
\label{eq:map}
\end{align}
We find an approximate solution to the optimization problem~\labelcref{eq:map}
using stochastic gradient ascent on the joint log probability,
according to the following procedure,
where $B \le N$ is a mini-batch size and $(\epsilon_1, \epsilon_2, \dots)$ is a
sequence of learning rates:
% \begin{listing}[h]
% \FrameSep0pt
% \begin{framed}
% \setlength{\abovedisplayskip}{0pt}
% \setlength{\belowdisplayskip}{0pt}
\begin{align}
&\mbox{Initialize } \theta_0 \sim \pi_f\pi_y;\; t \gets 0 \\
&\mbox{Repeat until convergence} \notag \\
&\quad\set{I_1, \dots, I_B} \sim \textrm{Uniform}(\set{K \subset [N] \mid \mathrm{card}({K})=B}) \\
&\quad\hat{g}_t = \begin{aligned}[t]
  \nabla_{\theta} \Big[
  \log \pi_f(\theta_f) + \log \pi_y(\theta_y)
  + \textstyle\frac{N}{B}
    \sum_{j=1}^B
      \log\left(\mathrm{Dist}(y(\br_{I_j}); F_{\theta_f}(\br_{I_j}), \theta_y)\right)
      \Big]_{\theta_{t-1}}
    \end{aligned}
    \label{eq:map-gradient}
    \\
&\quad\theta_t = \theta_{t-1} + \epsilon_t\hat{g}_t; \; t \gets t + 1.
\end{align}
% \end{framed}
% \caption{Stochastic MAP estimation for Bayesian Neural Field.}
% \label{listing:map}
% \end{listing}

We construct an overall ``deep ensemble'' $\set{(\theta_f^i, \theta_y^i)}_{i=1}^M$
containing $M \ge 1$ MAP estimates by repeating the above procedure $M$ times,
each with a different initialization of $\theta_0$ and random seed.

\paragraph{Stochastic Variational Inference}
A more uncertainty-aware alternative to MAP ensembles
is mean-field variational inference, which uses a
surrogate posterior $q_\phi(\theta) =
\prod_{i=1}^{n_f}\nu(\theta_{f,i}; \phi_{f,i})
\prod_{i=1}^{n_y}\nu(\theta_{y,i}; \phi_{y,i})
$ over $\Theta$ to approximate the true posterior
$P(\theta_f,\theta_y \mid \mathcal{D})$~\labelcref{eq:posterior-distribution}
given the data $\mathcal{D}$.
Optimal values for the variational parameters
$\phi = (\phi_{f,1},\dots,\phi_{f,n_f},\phi_{y,1},\dots,\phi_{y,n_y})$
are obtained by maximizing the ``evidence lower bound'':
\begin{align}
  \mathrm{ELBO}(\phi) &= \log P(\mathcal{D}) - \mathrm{KL}(q_\phi(\theta)\mid\mid P(\theta\mid\mathcal{D}))
  = \mathbb{E}_\phi\left[\log\frac{P(\mathcal{D}, \theta)}{q_\phi(\theta)}\right]
  \\
  &= \mathbb{E}_\phi[\log P(\mathcal{D}\mid\theta)] - \mathrm{KL}(q_\phi(\theta)\mid\mid \pi(\theta)).
  \label{eq:elbo}
\\
  &=\begin{aligned}[t]
    &\sum_{i=1}^N\mathbb{E}_{\phi}\left[
      \log\left(\mathrm{Dist}(y(\br_i); F_{\theta_f}(\br_i), \theta_y)\right)
      \right]
  \\
  &- \sum_{i=1}^{n_f}
      \mathbb{E}_{\phi_{f,i}}\left[
        \log\left(\frac{\nu(\theta_{f,i}; \phi_{f,i})}{\pi_f(\theta_{f,i})}\right)
        \right]
    -
    \sum_{i=1}^{n_y}
      \mathbb{E}_{\phi_{y,i}}\left[
        \log\left(\frac{\nu(\theta_{y,i}; \phi_{y,i})}{\pi_y(\theta_{y,i})}\right)
        \right].
      \hspace{-1cm}
  \end{aligned}
  \label{eq:elbo-full}
\end{align}
where \cref{eq:elbo-full} follows from the independence of the priors.
Finding the maximum of \cref{eq:elbo-full} is a challenging optimization problem.
Our implementation leverages a Gaussian variational posterior $q_\phi$
with KL reweighting, as described in \citet[Sec.~3.2, 3.4]{Blundell2015}.

Mean-field variational inference is known to underestimate posterior
variance and can also get stuck in local optima of \cref{eq:elbo}.
To alleviate these problems, we use a
\emph{variational ensemble} that is analogous to the MAP ensemble
described above.
More specifically, we first perform $M \ge 1$ runs of stochastic
variational inference with different initializations and random seeds,
which gives us an ensemble $\set{\phi^i, i = 1, \dots, M}$ of
variational parameters.
We then approximate the posterior $P(\theta \mid \mathcal{D})$ with an equal-weighted
mixture of the resulting variational distributions $\set{q_{\phi^i}}_{i=1}^M$.

\subsection*{Prediction Queries}
\label{sec:methods-queries}

We can approximate the posterior~\labelcref{eq:posterior-distribution}
using a set of samples $\set{(\theta_f^i, \theta_y^i)}_{i=1}^M$, which
may be obtained from either MAP ensemble estimation or stochastic variational
inference (by sampling from the ensemble of $M$ variational distributions).
We can then approximate the posterior-predictive distribution
$P(Y(\br_*)\mid\mathcal{D})$ (which marginalizes out the parameters
$\Theta$) of $Y(\br_*)$ at a novel field index $\br_* = (\s_*, t_*)$
by a mixture model with $M$ equally weighted components:
\begin{align}
\hat{P}(Y(\br_*) \mid \mathcal{D})
= \frac{1}{M} \sum_{i=1}^M
  \mathrm{Dist}(Y(\br_*); F_{\theta^i_f}(\br_*), \theta^i_y).
\label{eq:posterior-mixture}
\end{align}
Equipped with \cref{eq:posterior-mixture}, we can directly
compute predictive probabilities of events $\set{Y(\br_*) \le y}$,
predictive probability densities $\set{Y(\br_*) = y}$,
or conditional expectations
  $\mathbb{E}\left[ \varphi(Y(\br_*)) \mid \mathcal{D} \right]$
  for a probe function $\varphi: \mathbb{R} \to \mathbb{R}$.
Prediction intervals around $Y(\br_*)$ are estimated by computing
the $\alpha$-quantile $y_\alpha(\br_*)$, which satisfies
\begin{align}
\hat{P}(Y(\br_*) \le y_\alpha(\br_*) \mid \mathcal{D}) = \alpha
&& \alpha \in [0,1].
\label{eq:quantile}
\end{align}
For example, the median estimate is $y_{0.50}(\s_*,t_*)$ and 95\% prediction
interval is $[y_{0.025}(\s_*,t_*), y_{0.975}(\s_*,t_*)]$.
The quantiles~\labelcref{eq:quantile} are estimated numerically using
\citeauthor{Chandrupatla1997}'s
root finding algorithm \citep{Chandrupatla1997} on the cumulative distribution function of the
mixture~\labelcref{eq:posterior-mixture}.

\subsection*{Temporal Seasonal Features}
\label{sec:methods-periods}

Temporal seasonal features (c.f.~\cref{eq:cov-seasonal})
can enable more accurate prediction.
\Cref{table:periods} shows example periodic multiples $p$ for datasets with
various time units and seasonal components.
\begin{table}[h]
\caption{Examples of seasonal features for \bnf{}.
Each entry denotes the period $p$
corresponding to a measurement frequency and seasonal effect.
Given a period $p$ and time point $t$, the seasonal features are
$[\cos(2\pi h/pt), \sin(2\pi h/pt)]$ for harmonics $h=1,\dots,p/2$. }
\label{table:periods}
\begin{tabular*}{\textwidth}{c}\end{tabular*}
\begin{adjustbox}{max width=\textwidth}
\begin{tabular}{llrrrrrrrr}
\multicolumn{1}{c}{} & \multicolumn{1}{c}{}     & \multicolumn{8}{c}{\bfseries Seasonal Effect} \\ \cmidrule{2-10}
~ & ~       & Secondly & Minutely & Hourly & Daily   & Weekly   & Monthly   & Quarterly & Yearly \\ \cmidrule{2-10}
\multirow{8}*{\rotatebox{90}{\bfseries Measurements}}
  & Yearly    & -- & -- & --   & --    & --     & --      & --      & 1 \\
~ & Quarterly & -- & -- & --   & --    & --     & --      & 1       & 4 \\
~ & Monthly   & -- & -- & --   & --    & --     & 1       & 3       & 12 \\
~ & Weekly    & -- & -- & --   & --    & 1      & 4.35    & 13.045  & 52.18 \\
~ & Daily     & -- & -- & --   & 1     & 7      & 30.44   & 91.32   & 365.25 \\
~ & Hourly   & -- & -- & 1    & 24    & 168    & 730.5   & 2191.5  & 8766 \\
~ & Minutely  & -- & 1  & 60   & 1440  & 10080  & 43830   & 131490  & 525960 \\
~ & Secondly  & 1  & 60 & 3600 & 86400 & 604800 & 2629800 & 7889400 & 31557600 \\ \cmidrule{2-10}
\end{tabular}
\end{adjustbox}
\end{table}

\subsection*{Ablations}
\label{sec:methods-ablations}

To better understand how the prediction accuracy of \bnf{} varies
with the choices of inference algorithm and network architecture,
results from two classes of ablation studies for the benchmarks in
\cref{table:evaluation} are reported.

\paragraph{Inference Methods: Comparison of VI, MAP, and MLE}

\Cref{fig:runtime} shows a comparison of runtime vs.~accuracy profiles
on the six benchmarks from \cref{table:datasets} using three parameter
inference methods for \bnf{}---VI, MAP, and MLE.
MLE is the maximum likelihood estimation baseline described in
\citet{Lakshminarayanan2017}, which is identical to \cref{listing:bnf}
expect that the terms $\pi_f$ and $\pi_y$ in \cref{eq:map-gradient}
are ignored.
MLE performs no better than MAP or VI in all 18/18 profiles (and is
typically worse), illustrating the benefits of parameter priors and
posterior uncertainty which do not impose runtime overhead.
Between MAP and VI, the latter performs better in 13/18 profiles: that
is, on all metrics for Wind, Air Quality 1, and Air Quality 2; on RMSE
and MAE for Chickenpox; and on RMSE and MIS for Precipitation.

\afterpage{
\input{fig-runtime}
\input{fig-ablations}
\clearpage
}

\paragraph{Model Architectures}

\Cref{fig:ablations-1} shows the percentage change in RMSE, MAE, MIS,
and runtime using \bnf{} (MAP inference; 64 particles; fixed number of
training epochs) while applying a single change to the reference model
for each benchmark.
The goal of these ablations is to study how changes to the network
structure affect the predictive performance.

\cref{fig:ablations-depth-down,fig:ablations-depth-up} show
results for decreasing or increasing the network depth by one layer.
  The Sea Surface Temperature benchmark is the most sensitive to the
  network depth, where decreasing the depth causes the forecast errors
  to increase by around $50\%$, whereas increasing the depth delivers
  $5$--$10\%$ decreases.
  The MIS error is particularly sensitive to reducing the
  network depth where the results become significantly worse in 5/6
  benchmarks, although the runtime also decreases by up to $50\%$.

\cref{fig:ablations-width-down,fig:ablations-width-up} shows results
  for halving or doubling the width of the hidden layers.
  The Sea Surface Temperature benchmark is highly sensitive to halving
  the network width, with errors increasing above $25\%$.
  The remaining benchmarks demonstrate slight improvements in the errors
  which are not statistically significant, suggesting that the runtime gains
  could justify halving the width in these benchmarks.
  Doubling the with causes substantial increases in the runtime with no
  systematic pattern in the RMSE, MAE, or MIS values across the benchmarks.

\cref{fig:ablations-tanh,fig:ablations-elu} show results
  using only tanh or elu activations instead of the convex combination layer.
  Discarding the convex combination layer delivers runtime speedups, which
  are larger using tanh as compared to elu.
  However, there is no clear winner in terms of error when using only
  tanh or only elu; and no error metric is consistently negative by
  selecting one of the two activations.
  The changes in error which are consistently positive (as compared to
  the convex combination layer) are
  \begin{enumerate*}[label=(\roman*)]

    \item tanh only: Air Quality 2 (MIS $16\%$);

    \item elu only: Sea Surface Temperature (RMSE $59\%$, MAE
    $76\%$, MIS $49\%$) and Precipitation (MAE $7.8\%$, MIS $16\%$).
  \end{enumerate*}
  %
  % These results indicate the effectiveness of the convex
  % combination layers for learning activations.

\cref{fig:ablations-scaling} shows results for disabling the covariate
  scaling layer.
  The runtime is only slightly changed in all benchmarks. However,
  several errors increase consistently on average, namely in the
  Precipitation (RMSE $24\%$, MAE $27\%$, MIS $33\%$), Chickenpox (MIS
  $32\%$), and Air Quality 1 (MAE $13\%$) benchmarks.
  The remaining changes are neither consistently above nor
  below zero.
  %
  % The results support that the covariate scaling layer is highly
  % beneficial and does not introduce significant runtime overhead.

\cref{fig:ablations-fourier} shows results for omitting the spatial
  Fourier features (\cref{eq:cov-fourier}).
  While omitting these features delivers small runtime improvements,
  it also causes substantial increases in RMSE, MAE, and MIS values
  across all benchmarks except for Wind.
  These results support the hypothesis that spatial Fourier features
  are essential for accurate generalization across space and time.

In summary, the results (specifically
\crefrange{fig:ablations-tanh}{fig:ablations-fourier}), demonstrate
that architectural choices in \bnf{} such as the spatial Fourier
features, convex combination layer, and covariate scaling
are effective in reducing the prediction error across several benchmarks
and metrics at the cost of a manageable runtime overhead.

\subsection*{Evaluation Metrics}

The quality of point forecasts are evaluated using RMSE and MAE scores.
Interval forecasts are evaluated using the MIS score at level $\alpha =0.05$.
The definitions are as follows:
\begin{align}
&\mbox{Root Mean Squared Error (RMSE)}
  &&\sqrt{\textstyle\sum_{i=1}^n (y_i - \hat{y}_i)^2/n} \label{eq:rmse}\\
&\mbox{Mean Absolute Error (MAE)}
  &&\textstyle\sum_{i=1}^{n} \abs{y_i - \hat{y}_i}/n \label{eq:mae}\\
& \mbox{Mean Interval Score (MIS)}
  &&\begin{aligned}[t]
    \textstyle\sum_{i=1}^{n}
    \Big[
    &(u_i - \ell_i)
    +  \frac{2}{\alpha}(\ell_i-y_i)\mathbf{1}[y_i < \ell_i] \label{eq:mis}\\
    &\;+  \frac{2}{\alpha}(y_i < u_i)\mathbf{1}[u_i < y_i]
    \Big]\Big/n,
  \end{aligned}
\end{align}
where $y_i$ is the true value, $\hat{y}_i$ is the point forecast, and
$(\ell_i, u_i)$ are endpoints of the interval forecast.

\backmatter

\section*{Data Availability}

All datasets from \cref{table:datasets} are publicly available under
open-source licenses.
\begin{itemize}
\item Wind Speed. GNU GPL v2.
\url{https://r-spatial.github.io/gstat/reference/wind.html}.

\item Air Quality 1. GNU GPL v3.
\url{https://rdrr.io/cran/spacetime/man/air.html}.

\item Air Quality 2. CC Attribution 1.0 Generic.
\url{https://doi.org/10.5281/zenodo.4531304}.

\item Chickenpox Cases. CC Attribution 4.0 International.
\url{https://doi.org/10.24432/C5103B}.

\item Precipitation. Public Domain.
\url{https://www.image.ucar.edu/Data/US.monthly.met/}.

\item Sea Surface Temperature. GNU GPL v2.
\url{https://github.com/andrewzm/STRbook/}.
\end{itemize}

The full datasets, test/train splits, model predictions, and ablation
results are available at \url{https://doi.org/10.5281/zenodo.12735404}.
Refer to the README in these files for additional information.

\section*{Code Availability}
An open-source Python implementation of \bnf{} is available at
\url{https://github.com/google/bayesnf}
under an Apache-2.0 License.
The full evaluation pipeline containing all model configurations for
the baselines is also provided in the repository.

\bibliography{paper}

\section*{Author Contributions}

The statistical model was designed and implemented by F.S, M.H, J.B., and
U.K. Evaluations were designed by F.S. and implemented by F.S., J.B, C.C,
and BP. R.S. and B.P provided guidance and oversight. F.S.\ drafted the
manuscript, all authors contributed to its revision and completion.

\section*{Competing Interests}

The authors declare no competing interests.

\clearpage

\listoftables

\listoflistings

\listoffigures

\end{document}

%% file: fig-network-vertical.tex
%!TEX root=./paper.tex

\begin{figure}[p]%
\centering

\makeatletter
\DeclareRobustCommand{\rvdots}{%
  \vbox{
    \baselineskip2\p@\lineskiplimit\z@
    \kern-\p@
    \hbox{.}\hbox{.}\hbox{.}
  }}
\makeatother

\begin{tikzpicture}
\tikzstyle{unit}=[draw=black,circle,fill=white]
\tikzstyle{observable}=[fill=black!20!white]
\tikzstyle{param}=[fill=red!20!white]
\tikzstyle{connection}=[color=black,line width=.1px]
\tikzstyle{layerLabel}=[font=\normalsize]
\tikzstyle{nodeLabel}=[inner xsep=2pt]
\footnotesize

% SpatioTemporal Coordinates
\node[name=s2,unit,observable,at={(0,0)},,label={above:$\cdots$}]{};
\node[name=s1,unit,observable,left=.1cm of s2,label={above:$s_2$}]{};
\node[name=s0,unit,observable,left=.1cm of s1,label={above:$s_1$}]{};

\node[name=s3,unit,observable,right=.1cm of s2,label={above:$s_d$}]{};
\node[name=s4,unit,observable,right=.1cm of s3,label={[name=tlabel]above:$t$}]{};

\draw [decoration={brace,raise=0.4cm},decorate]
  (s0.north) -- node[name=slabel,pos=0.5,yshift=.65cm]{$\s$} (s3.north);

% Covariate Layer
\foreach \i in {0,...,4} { \node[name=y\i,unit,observable,below=.3 of s\i]{}; }
\node[name=y5,unit,observable,right=.1 of y4]{};
\node[name=y6,unit,observable,right=.1 of y5,label={[nodeLabel]right:$x_m(\s{,}t)$}]{};
\node[name=y7,unit,observable,left=.1 of y0]{};
\node[name=y8,unit,observable,left=.1 of y7,label={[nodeLabel]left:$x_1(\s{,}t)$}]{};]{};
\begin{scope}[on background layer]
\foreach \i in {0,...,4}{
    \foreach \j in {0,...,8}{
    \draw[connection] (s\i.south) -- (y\j.north);
    }
}
\end{scope}

% Covariate Layer
\foreach \i in {0,1,2,3,4,5,7} { \node[name=x\i,unit,below=.6 of y\i]{}; }
\node[name=x6,unit,right=.1 of x5,label={[nodeLabel]right:$h^0_m(\s{,}t)$}]{};
\node[name=x8,unit,left=.1 of x7,label={[nodeLabel]left:$h^0_1(\s{,}t)$}]{};]{};
\begin{scope}[on background layer]
\foreach \j in {0,...,8}{ \draw[connection, line width=.2px] (y\j.south) -- (x\j.north); }
\end{scope}

% Linear Layer 1
\foreach \i in {0,...,8} {
    \node[name=z1-\i,unit,below=.85 of x\i]{};
}
\node[name=z1-9,unit,left=.1 of z1-8,label={[nodeLabel]left:$z^1_{1}(\s{,}t)$}]{};
\node[name=z1-10,unit,right=.1 of z1-6,label={[nodeLabel]right:$z^1_{N_1}(\s{,}t)$}]{};
\foreach \i in {0,...,8}{
    \foreach \j in {0,...,10}{
        \begin{scope}[on background layer]
        \draw[connection] (x\i.south) -- (z1-\j.north);
        \end{scope}
    }
}
\node[name=W1-center,param,fill=none,rectangle,at={($(x2)!0.5!(z1-2)$)},anchor=center] {};

% Activation Layer 1
\node[name=z1elu-0,unit,inner sep=1.5pt,below right=.85 and 2 of z1-2,anchor=center]{};
\node[name=z1tan-10,unit,inner sep=1.5pt,below left=.85 and 2 of z1-2,anchor=center]{};
\foreach \i in {1,...,10}{
    \pgfmathsetmacro{\eprev}{int(\i-1)}
    \pgfmathsetmacro{\itan}{int(10-\i)}
    \pgfmathsetmacro{\itanprev}{int(\itan+1)}
    \node[name=z1elu-\i,unit,inner sep=1.5pt,left = .1 of z1elu-\eprev,anchor=center]{};
    \node[name=z1tan-\itan,unit,inner sep=1.5pt,right = .1 of z1tan-\itanprev,anchor=center]{};
}

\foreach \i/\l/\direction in {elu/elu/right, tan/tanh/left} {
    \node[
        name=\i,
        thick,
        draw=black, fill=none,
        inner sep=2.5pt,
        outer sep=0pt, fit=(z1\i-0) (z1\i-10),
        label={[font=\tiny,inner ysep=1pt]\direction:\l}
        ] {};
}

% Hidden Layer 1
\foreach \i in {0,...,10} { \node[name=h1-\i,unit,below=2*.85 of z1-\i]{}; }
\node[at={(h1-9)}, label={[nodeLabel]left:$h^1_{1}(\s{,}t)$}]{};
\node[at={(h1-10)}, label={[nodeLabel]right:$h^1_{N_1}(\s{,}t)$}]{};

\node[name=gamma1, unit, param, right = 1 of z1elu-0, label={right:$\gamma^1$}]{};
\begin{scope}[on background layer]
\foreach \i in {0, ..., 10} {
    \draw[connection](gamma1.south) -- (h1-\i.north);
}
\end{scope}

\node[name=W1,
  unit,param,
  at = {(gamma1 |- x2)},
  anchor=center,
  label={[name=W1-label]right:$\omega^{1}_{ij}$}
  ]{};
\node[name=W1-bias,
  unit,param,
  below=.35 of W1,
  anchor=center,
  label={[name=W1-bias-label]right:$\beta^{1}_{i}$}
  ]{};
\node[
    name=W1-plate,
    draw=black, fill=none,
    inner sep=2.5pt,
    outer sep=0pt,
    fit={(W1.north west) ([xshift=.5cm]W1-bias.south east)},
    label={[name=W1-index,inner sep=2pt,font=\tiny]below:
      $\begin{aligned}
      i&{\in}\,[N_1] \\[-2.5pt]
      j&{\in}\,[m]
      \end{aligned}$
      },
      ]{};
\node[name=W1-scale,
    unit,param,
    above=.25 of W1,
    label={right:$\xi^1$}] {};
\draw[-stealth] (W1-scale) -- (W1);
\draw[-stealth] (W1-scale) to[bend left] (W1-bias);

\begin{scope}[on background layer]
\foreach \j in {0,...,10} {
    \draw[connection] (W1.south) -- (z1-\j.west);
    \draw[connection] (W1-bias.south) -- (z1-\j.west);
    }
\end{scope}

\node[name=W0-scale,
    unit,param,
    above=.25 of W1-scale,
    draw=black,
    label={right:$\xi^0_i$}] {};
\node[
    name=W0-plate,
    draw=black, fill=none,
    inner sep=2.5pt,
    outer sep=0pt,
    fit={(W0-scale.north west) ([xshift=.5cm]W0-scale.south east)},
    label={[name=W0-index,inner sep=2pt,font=\tiny]above:
      $i{\in}[m]$
      },
    % label={[label distance=10pt]above:$\bm\theta_f$},
    ]{};

\begin{scope}[on background layer]
\foreach \i in {0, ..., 8} {
    \draw[connection] (W0-scale) -- (x\i);
}
\end{scope}

\begin{scope}[on background layer]
\foreach \i in {elu, tan} {
\foreach \prefix/\e/\w in {z1/south/north, h1/north/south} {
    \draw[connection](\prefix-9.\e) -- (z1\i-10.\w);
    \draw[connection](\prefix-10.\e) -- (z1\i-0.\w);
    \draw[connection](\prefix-6.\e) -- (z1\i-1.\w);
    \draw[connection](\prefix-5.\e) -- (z1\i-2.\w);

    \draw[connection](\prefix-0.\e) -- (z1\i-7.\w);
    \draw[connection](\prefix-1.\e) -- (z1\i-6.\w);
    \draw[connection](\prefix-2.\e) -- (z1\i-5.\w);
    \draw[connection](\prefix-3.\e) -- (z1\i-4.\w);
    \draw[connection](\prefix-4.\e) -- (z1\i-3.\w);

    \draw[connection](\prefix-7.\e) -- (z1\i-8.\w);
    \draw[connection](\prefix-8.\e) -- (z1\i-9.\w);
    }
}
\end{scope}

% Linear Layer 2
\foreach \i in {0,...,10} {
    \node[name=z2-\i,unit,below=.85 of h1-\i]{};
}
\foreach \i in {0,...,10}{
    \foreach \j in {0,...,10}{
    \begin{scope}[on background layer]
    \draw[connection] (h1-\i.south) -- (z2-\j.north);
    \end{scope}
    }
}
\node[at={(z2-9)}, label={[nodeLabel]left:$z^2_1(\s{,}t)$}]{};
\node[at={(z2-10)}, label={[nodeLabel]right:$z^2_{N_2}(\s{,}t)$}]{};

\node[name=W2,
  unit,param,
  at = {(gamma1 |- h1-2)},
  anchor=center,
  label={[name=W2-label,inner sep=0pt,xshift=.05cm]right:$\omega^{2}_{ij}$}
  ]{};
\node[name=W2-bias,
  unit,param,
  below=.35 of W2,
  anchor=center,
  label={[name=W2-bias-label,inner sep=0pt,xshift=.05cm]right:$\beta^{2}_{i}$}
  ]{};
\node[
    name=W2-plate,
    draw=black, fill=none,
    inner sep=2.5pt,
    outer sep=0pt,
    fit={(W2.north west) ([xshift=.5cm]W2-bias.south east)},
    label={[name=W2-index,inner sep=2pt,font=\tiny]below:
      $\begin{aligned}
      i&{\in}\,[N_2] \\[-2.5pt]
      j&{\in}\,[N_1]
      \end{aligned}$
      },
      ]{};
\node[name=W2-scale,
    unit,param,
    above=.25 of W2,
    label={right:$\xi^2$}] {};
\draw[-stealth] (W2-scale) -- (W2);
\draw[-stealth] (W2-scale) to[bend left] (W2-bias);

\begin{scope}[on background layer]
\foreach \j in {0,...,10} {
    \draw[connection] (W2.south) -- (z2-\j.north);
    \draw[connection] (W2-bias.south) -- (z2-\j.north);
    }
\end{scope}

% Activation Layer 2.
\node[name=z2elu-0,unit,inner sep=1.5pt,below right=.85 and 2 of z2-2,anchor=center]{};
\node[name=z2tan-10,unit,inner sep=1.5pt,below left=.85 and 2 of z2-2,anchor=center]{};
\foreach \i in {1,...,10}{
    \pgfmathsetmacro{\eprev}{int(\i-1)}
    \pgfmathsetmacro{\itan}{int(10-\i)}
    \pgfmathsetmacro{\itanprev}{int(\itan+1)}
    \node[name=z2elu-\i,unit,inner sep=1.5pt,left = .1 of z2elu-\eprev,anchor=center]{};
    \node[name=z2tan-\itan,unit,inner sep=1.5pt,right = .1 of z2tan-\itanprev,anchor=center]{};
}

\foreach \i/\l/\direction in {elu/elu/right, tan/tanh/left} {
    \node[
        name=\i,
        thick,
        draw=black, fill=none,
        inner sep=2.5pt,
        outer sep=0pt, fit=(z2\i-0) (z2\i-10),
        label={[font=\tiny,inner ysep=1pt]\direction:\l}
        ] {};
}

% Hidden Layer 2.
\foreach \i in {0,...,10} { \node[name=h2-\i,unit,below=2*.85 of z2-\i]{}; }
\node[at={(h2-9)}, label={[nodeLabel]left:$h^2_{1}(\s{,}t)$}]{};
\node[at={(h2-10)}, label={[nodeLabel]right:$h^2_{N_2}(\s{,}t)$}]{};

\node[name=gamma2, unit, param, at={(gamma1 |- z2elu-0)}, label={right:$\gamma^2$}]{};
\begin{scope}[on background layer]
\foreach \i in {0, ..., 10} {
    \draw[connection](gamma2.south) -- (h2-\i.north);
}
\end{scope}

\begin{scope}[on background layer]
\foreach \i in {elu, tan} {
\foreach \prefix/\e/\w in {z2/south/north, h2/north/south} {
    \draw[connection](\prefix-9.\e) -- (z2\i-10.\w);
    \draw[connection](\prefix-10.\e) -- (z2\i-0.\w);
    \draw[connection](\prefix-6.\e) -- (z2\i-1.\w);
    \draw[connection](\prefix-5.\e) -- (z2\i-2.\w);

    \draw[connection](\prefix-0.\e) -- (z2\i-7.\w);
    \draw[connection](\prefix-1.\e) -- (z2\i-6.\w);
    \draw[connection](\prefix-2.\e) -- (z2\i-5.\w);
    \draw[connection](\prefix-3.\e) -- (z2\i-4.\w);
    \draw[connection](\prefix-4.\e) -- (z2\i-3.\w);

    \draw[connection](\prefix-7.\e) -- (z2\i-8.\w);
    \draw[connection](\prefix-8.\e) -- (z2\i-9.\w);
    }
}
\end{scope}

% Output Layer
\node[name=output, unit, below=.55 of h2-2,label={left:$z^3_1(\s,t)\equiv F(\s{,}t)$}]{};
\begin{scope}[on background layer]
\foreach \i in {0,...,10} {
    \draw[connection](h2-\i.south) -- (output.north);
}
\end{scope}

\node[name=W3,
  unit,param,
  at = {(gamma1 |- h2-2)},
  anchor=center,
  label={[name=W3-label,inner sep=0pt,xshift=.05cm]right:$\omega^{3}_{ij}$}
  ]{};
\node[name=W3-bias,
  unit,param,
  below=.35 of W3,
  anchor=center,
  label={[name=W3-bias-label,inner sep=0pt,xshift=.05cm]right:$\beta^{3}_{i}$}
  ]{};
\node[
    name=W3-plate,
    draw=black, fill=none,
    inner sep=2.5pt,
    outer sep=0pt,
    fit={(W3.north west) ([xshift=.5cm]W3-bias.south east)},
    label={[name=W3-index,inner sep=2pt,font=\tiny]below:
      $\begin{aligned}
      i&{\in}\,[N_3] \\[-2.5pt]
      j&{\in}\,[N_2]
      \end{aligned}$
      },
      ]{};
\node[name=W3-scale,
    unit,param,
    above=.25 of W3,
    label={right:$\xi^3$}] {};
\draw[-stealth] (W3-scale) -- (W3);
\draw[-stealth] (W3-scale) to[bend left] (W3-bias);

\begin{scope}[on background layer]
\draw[connection] (W3.south) -- (output.north);
\draw[connection] (W3-bias.south) -- (output.north);
\end{scope}

% % Observation Layer
\node[name=field, unit, observable, below=.4 of output,label={left:$Y(\s{,}t)$}]{};
\node[name=phi, unit, param, at={(gamma2|-field)},label={right:$\theta_y$}]{};
\draw[connection,-stealth](phi) -- (field);
\draw[connection,-stealth](output.south) -- (field.north);

% \node[name=layer-output,at={(layer-st -| field)},

% Layer names
\node[layerLabel,name=layer-st,left=2.5 of s0.west,anchor=east]{\begin{tabular}{@{}c@{}}Spatiotemporal Coordinates\end{tabular}};
\node[name=aux,at={(y1)}]{};
\node[layerLabel,at={(aux -| layer-st.west)},anchor=west]{\begin{tabular}{@{}c@{}}Spatiotemporal Covariates\end{tabular}};
\node[name=aux,at={($(y2)!0.75!(x2)$)}]{};
\node[name=label-scale,layerLabel,at={(aux -| layer-st.west)},anchor=west]{\begin{tabular}{@{}c@{}}Covariate Scaling Layer\end{tabular}};
\node[name=aux,at={($(x2)!0.5!(z1-2)$)}]{};
\node[name=label-start-1,layerLabel,at={(aux -| layer-st.west)},anchor=west]{\begin{tabular}{@{}c@{}}Dense Linear Layer\end{tabular}};
\node[name=aux,at={($(z1-2)!0.5!($(z1-2)!0.5!(h1-2)$)$)}]{};
\node[layerLabel,at={(aux -| layer-st.west)},anchor=west]{\begin{tabular}{@{}c@{}}Nonlinear Activation Layer\end{tabular}};
\node[name=aux,at={($(h1-2)!0.5!($(z1-2)!0.5!(h1-2)$)$)}]{};
\node[name=label-end-1,layerLabel,at={(aux -| layer-st.west)},anchor=west]{\begin{tabular}{@{}c@{}}Convex Combination Layer\end{tabular}};
\node[name=aux,at={($(h1-2)!0.5!(z2-2)$)}]{};
\node[name=label-start-2,layerLabel,at={(aux -| layer-st.west)},anchor=west]{\begin{tabular}{@{}c@{}}Dense Linear Layer\end{tabular}};
\node[name=aux,at={($(z2-2)!0.5!($(z2-2)!0.5!(h2-2)$)$)}]{};
\node[layerLabel,at={(aux -| layer-st.west)},anchor=west]{\begin{tabular}{@{}c@{}}Nonlinear Activation Layer\end{tabular}};
\node[name=aux,at={($(h2-2)!0.5!($(z2-2)!0.5!(h2-2)$)$)}]{};
\node[name=label-end-2,layerLabel,at={(aux -| layer-st.west)},anchor=west]{\begin{tabular}{@{}c@{}}Convex Combination Layer\end{tabular}};
\node[name=aux,at={($(h2-2)!0.5!(output)$)}]{};
\node[name=label-output-start,layerLabel,at={(aux -| layer-st.west)},anchor=west]{\begin{tabular}{@{}c@{}}Dense Linear Layer\end{tabular}};
\node[name=aux,at={($(output)!0.5!(field)$)}]{};
\node[name=label-output-end,layerLabel,at={(aux -| layer-st.west)},anchor=west]{\begin{tabular}{@{}c@{}}Observation Layer\end{tabular}};

\foreach \i in {1, 2}{
    \draw[
        decorate,
        decoration={brace,amplitude=7.5pt}]
        (label-end-\i.south west)
        -- node[pos=0.5,label={[rotate=90,anchor=south,yshift=.25cm]left:Hidden Layer \i}]{}
        (label-start-\i.north west);
    }

\draw[
    decorate,
    decoration={brace,amplitude=7.5pt}]
    (label-scale.south west)
    -- node[pos=0.5,label={[rotate=90,anchor=south,yshift=.25cm]left:Input Layer}]{}
    (layer-st.north west);

\draw[
    decorate,
    decoration={brace,amplitude=7.5pt}]
    (label-output-end.south west)
    -- node[pos=0.5,label={[rotate=90,anchor=south,yshift=.25cm]left:Output Layer}]{}
    (label-output-start.north west);

\draw[
    decorate,
    decoration={brace,amplitude=7.5pt,raise=2pt}]
    (W0-plate.north east)
    -- node[pos=0.5,label={[xshift=5pt]right:$\theta_f$}]{}
    (W3-plate.south east);

% NYC Map
\node[
  name=nyc,
  above=.75 cm of s2,
  anchor=south,
  % at={($(s2)!0.5!(output)$)},
  %yshift=5.5cm,
  % label={above:longitude $(s_1)$},
  % label={[anchor=center,xshift=-.25cm,rotate=90]left:latitude $(s_2)$},
  % label={[inner sep=0pt]300:time $(t)$},
  ]
  {\includegraphics[scale=.55]{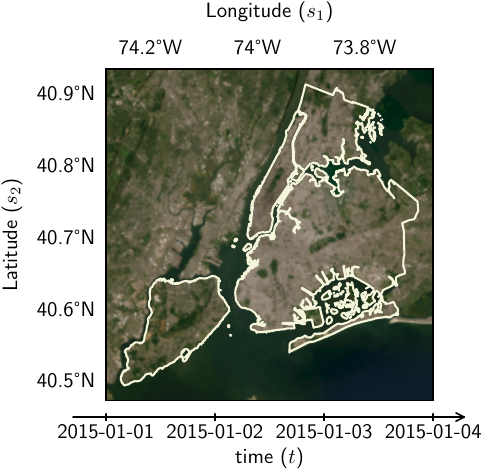}};
% \draw[-latex] (nyc.west) to[bend right=80] (slabel);
% \draw[-latex] (nyc.south) to[bend left] (tlabel);

\node[name=nyc-pred,
  below left=.25 and 2 cm of field,
  anchor=north]{
  \includegraphics[scale=.55]{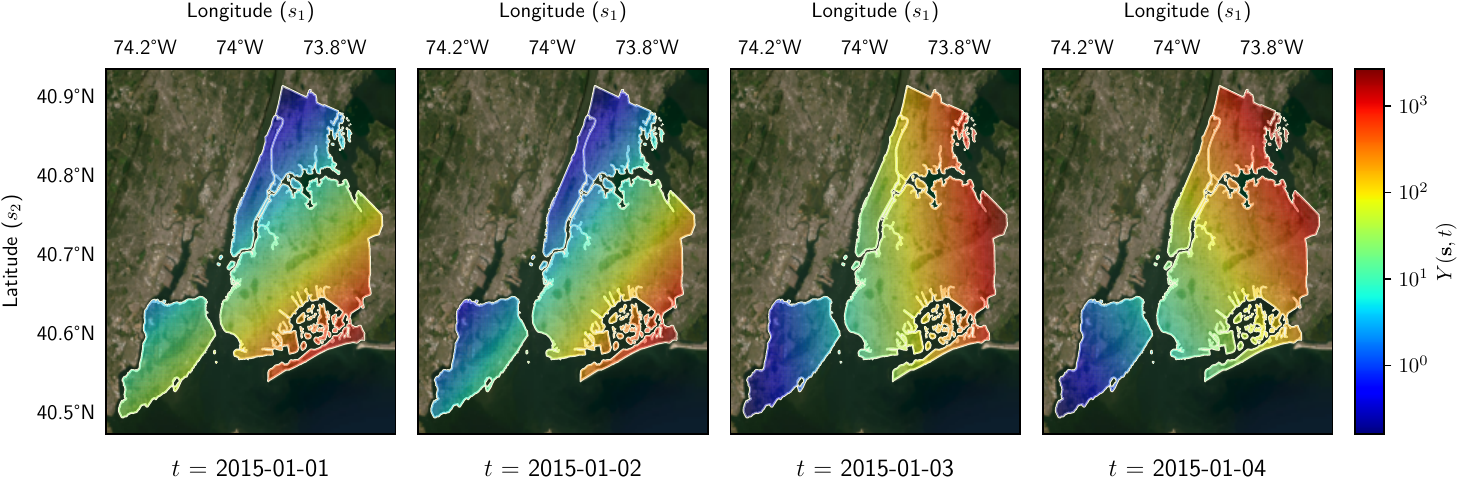}};

% Add legends.
\node[name=capt-a,text width=1cm,at=(nyc-pred.west |- nyc), anchor=west,draw=none] {
  \subcaption{}
  \label{fig:network-input}
  };
\node[name=capt-b,text width=1cm,at=(nyc-pred.west |- nyc.south), anchor=west,draw=none] {
  \subcaption{}
  \label{fig:network-structure}
  };
\node[name=capt-c,text width=1cm,at=(nyc-pred.west |- nyc-pred.north west), yshift=-.25cm, anchor=west,draw=none] {
  \subcaption{}
  \label{fig:network-output}
  };

\end{tikzpicture}
\caption{
Probabilistic graphical model representation of the Bayesian Neural Field.
\subref{fig:network-input} An example spatiotemporal domain comprised of
two spatial coordinates (latitude, longitude) and a daily time coordinate.
\subref{fig:network-structure} In the probabilistic graphical model,
each node denotes a model variable and each edge denotes a
direct relationship between a pair of variables. Gray
nodes are observed variables and white notes are local latent
variables, which are both associated with an observation
$Y(\s,t)$ at a spatiotemporal coordinate $(\s,t)$. Pink nodes are global
latent variables (parameters), which are shared across all spatiotemporal coordinates.
\subref{fig:network-output} Realizations of the spatiotemporal field generated
from the \bnf{} at four example time points.
Satellite basemap source: \citet{esri}.
}
\label{fig:network}
\end{figure}

%% file: fig-generative.tex
%!TEX root=./paper.tex
\begin{listing}[t]
\begin{framed}
\setlength{\abovedisplayskip}{0pt}
\setlength{\belowdisplayskip}{0pt}
\begin{align*}
&\mbox{\underline{\textbf{Configuration}}} \left\lbrace\begin{array}{ll}
  L&:\mbox{number of internal layers}\\
  x_i&: \mbox{covariate functions}\ (1 \le i \le m)\\
  N^\ell&: \mbox{number of hidden units}\ (1 \le \ell \le L)\\
  A^\ell&: \mbox{number of activation functions}\ (1 \le \ell \le L)\\
  u^\ell_j&: \mbox{activation functions}\ (1 \le \ell \le L; 1 \le j \le A^\ell) \\
  \mathrm{Dist}&: \mbox{observation distribution}
\end{array}\right.
\span\span\span\span\\
% (w^0_1, \dots, w^0_m) &\sim N(0, \ln(1+e^{\xi^{0}})) && i \in [N^\ell] \\
&\mbox{\underline{\textbf{Covariate Scaling Layer}}} \\
&(\xi^0_1, \dots, \xi^0_m) \simiid \mathrm{Normal}(0,1)
&& h^0(\s, t) \coloneqq (e^{\xi^0_1} x_1(\s,t), \dots, e^{\xi^0_m} x_m(\s,t)) \\
&\mbox{\underline{\textbf{Hidden Layers}} $\left(\ell=1,\dots,L+1; i=1,\dots,N^\ell\right)$} \span\span \\
&(\xi^{\ell}, \gamma^{\ell}_1, \dots, \gamma^{\ell}_{A^\ell}) \simiid \mathrm{Normal}(0,1),
&&z^{\ell}_{i}(\s,t) \coloneqq \textstyle\sum_{j=1}^{N^{\ell-1}} \frac{\omega^{\ell}_{ij}}{\sqrt{N^{\ell-1}}}h^{\ell-1}_j(\s,t) + \beta^{\ell}_i && \\
&\begin{aligned}[t]
(\omega^{\ell}_{i1}, \dots, \omega^{\ell}_{iN^{\ell-1}}, \beta^{\ell}_i) \simiid \mathrm{Normal}(0, \sigma^\ell) &\\
  \mbox{where } \sigma^\ell \coloneqq \ln(1+e^{\xi^{\ell}})
  \end{aligned}
\qquad
&&\begin{aligned}[t]
  h^{\ell}_{i}(\s,t) \coloneqq \textstyle\sum_{j=1}^{A^\ell}
    \frac{e^{\gamma^{\ell}_{j}}}{\sum_{k=1}^{A^\ell}{e^{\gamma^{\ell}_{k}}}}
    u^{\ell}_{j}(z^{\ell}_{i}(\s,t))&\\
  \mbox{(only if $\ell < L+1$)}&
\end{aligned}
\\
% &\mbox{\textbf{Hidden Output Layer}} \span\span \\
% &\xi^{H+1} \simiid \mathrm{Normal}(0,1), \, \sigma_{H+1} = \ln(1+e^{\xi^{H+1}}) \\
% &(\omega^{H+1}_{1:N^H}, \dots, \omega^{H+1}_{N^H}, \beta^{H+1}) \simiid N(0, \sigma^2_{H+1}/{N^H})
% &&F(\s,t) = \textstyle\sum_{i=1}^{N^H}\omega^{H+1}_j h^H_j(\s,t) + \beta^{H+1} \\
&\mbox{\underline{\textbf{Observation Layer}}} \span\span \\
&\Theta_y \sim \pi_y &&F(\s,t) \coloneqq z^{L+1}_1(\s,t) \\
&~ &&Y(\s, t) \sim \mathrm{Dist}(F(\s,t), \Theta_y)
\end{align*}
\end{framed}
\caption{Generative process for the Bayesian Neural Field in \cref{fig:network}.
Terms on the left are global parameters. Terms on the right
are local latent variables associated with a given spatiotemporal index $(\s,t)$.
}
\label{listing:bnf}
\end{listing}

%% file: fig-datasets.tex
%!TEX root=./paper.tex
% \begin{minipage}[c][\textheight][c]{\textwidth}
\begin{figure}[p]
\captionsetup{type=table}
\caption{Spatiotemporal datasets analyzed in the empirical evaluation.}
\label{table:datasets}
\begin{tabular*}{\textwidth}{c} \end{tabular*}
\begin{adjustbox}{max width=\textwidth,center}
\begin{tabular}{@{}lllrrrrll}
\toprule
\textbf{Dataset}                          & \textbf{Region} & \textbf{Frequency} & \textbf{Locations} & \textbf{Time Points} & \textbf{Observations} & \textbf{Missing} & \textbf{Start} & \textbf{End} \\ \midrule
Wind Speed~\citep{Haslett1989}            & Ireland         & Daily              & 12                 & 6574                 & 78,888                & 0\%              & 1961-01-01     & 1978-12-31 \\
Air Quality 1~\citep{Pebesma2012}         & Germany         & Daily              & 70                 & 4383                 & 149,151               & 52\%             & 1998-01-01     & 2009-12-31 \\
Air Quality 2~\citep{Hamelijnck2021}      & London          & Hourly             & 72                 & 2159                 & 144,570               & 7\%              & 2018-12-31     & 2019-03-31 \\
Chickenpox Cases~\citep{Chickenpox2021}   & Hungary         & Weekly             & 20                 & 522                  & 10,440                & 0\%              & 2005-01-03     & 2014-12-29 \\
Precipitation~\citep{Precipitation2010}   & Colorado        & Monthly            & 358                & 576                  & 134,800               & 35\%             & 1950-01-01     & 1997-12-01 \\
Sea Surface Temperature~\citep{Wikle2019} & Pacific Ocean   & Monthly            & 2261               & 399                  & 902,139               & 0\%              & 1970-01-01     & 2003-03-01 \\ \bottomrule
\end{tabular}
\end{adjustbox}

\captionsetup{type=figure}
\begin{subfigure}{\linewidth}
\centering
\includegraphics[width=.875\textwidth]{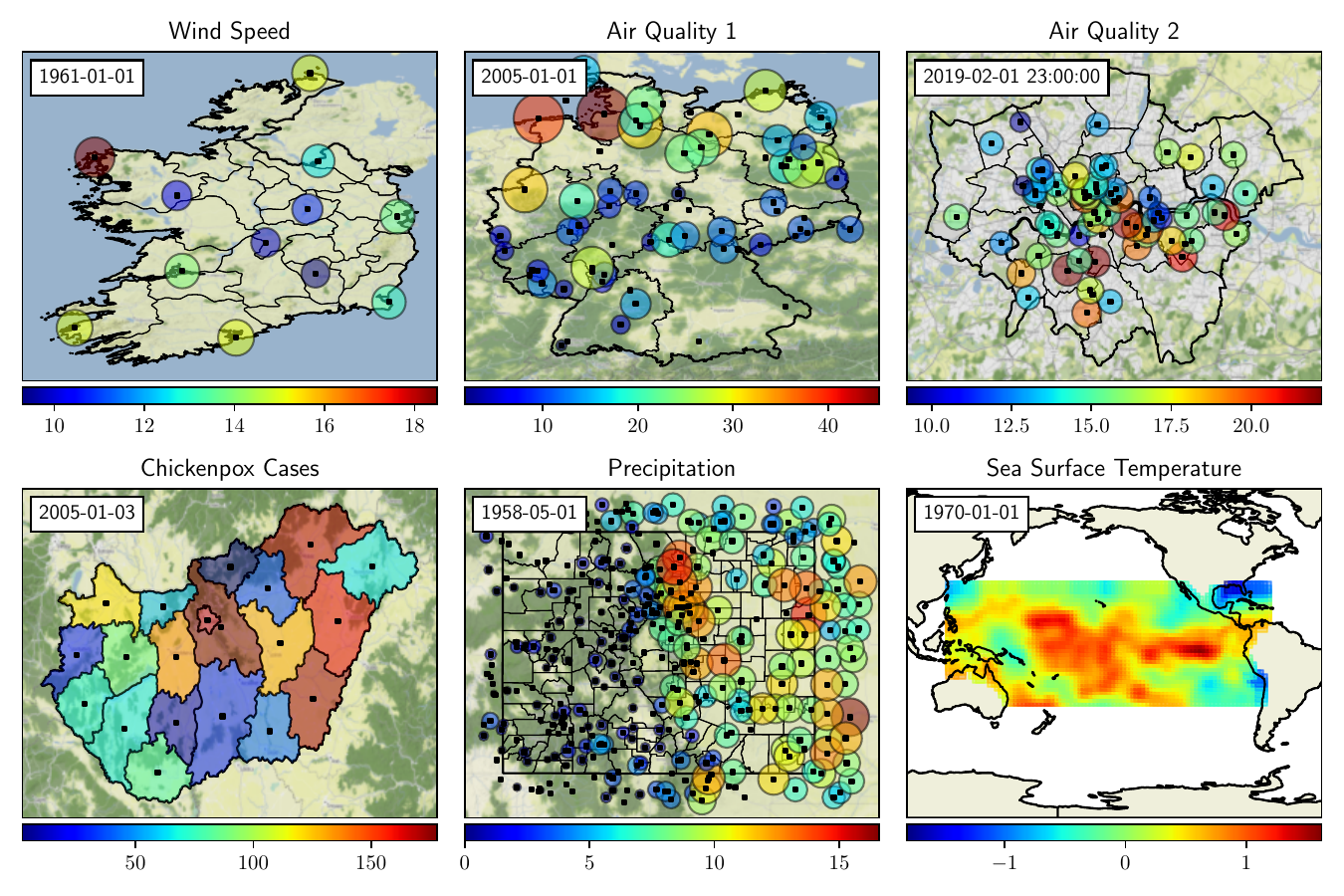}
\captionsetup{skip=0pt}
\caption{}
\label{fig:datasets-space}
\end{subfigure}
\begin{subfigure}{\linewidth}
\centering
\includegraphics[width=.875\textwidth]{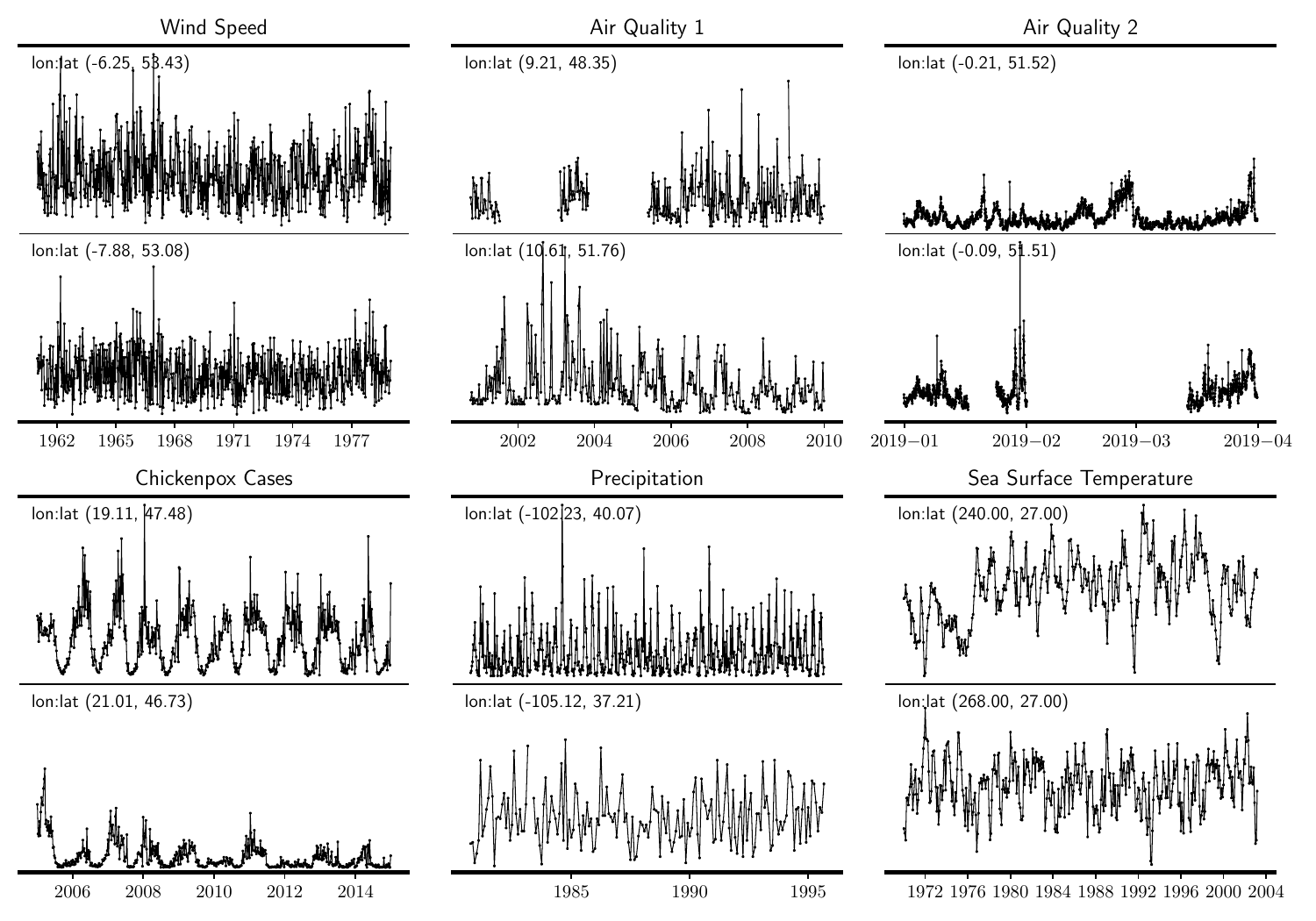}
\caption{}
\label{fig:datasets-time}
\end{subfigure}
\captionsetup{skip=0pt}
\caption{Spatial and temporal observations for evaluation datasets from \cref{table:datasets}.
\subref{fig:datasets-space} Snapshots of spatial observations at fixed points in time.
\subref{fig:datasets-time} Snapshots of temporal observations at fixed locations in space.
Satellite basemap source: \citet{stamen}.
}
\label{fig:datasets}
\end{figure}

%% file: fig-evaluation.tex
%!TEX root=./paper.tex
\begin{table}[p]
\caption{
Point prediction errors
in terms of root-mean square error (RMSE) and mean average error (MAE);
interval prediction error in terms of mean interval score (MIS);
and wall-clock runtime in seconds on spatiotemporal benchmark
datasets using multiple baselines methods.
Each error measurement, shown to two significant figures,
is an average over five independent test/train splits.
The symbol $\xmark$ denotes an experiment that failed to complete
successfully (timeout, out-of-memory, too sparse, etc.).
Bold values indicate statistically significant lowest errors
(Mann-Whitney U-Test at the 5\% level with Bonferroni correction).
}
\label{table:evaluation}
\newcommand{\basBNF}{BNF}
\newcommand{\basBNFVI}{BNF VI}
\newcommand{\basBNFMAP}{BNF MAP}
\newcommand{\basBNFNS}{BNF NS}
\newcommand{\basSTSVGP}{ST SVGP}
\newcommand{\basTSREG}{TSREG OLS}
\newcommand{\basSPDEAR}{ST GLMM AR1}
\newcommand{\basSPDERW}{ST GLMM RW}
\newcommand{\basSPDEIID}{ST GLMM IID}
\newcommand{\basSPDE}{ST GLMM (All)}
\newcommand{\basGBOOST}{ST GBOOST}
\newcommand{\basNBEATS}{NBEATS}
\renewcommand{\basBNF}{Bayesian Neural Field}
\renewcommand{\basBNFVI}{Bayesian Neural Field (VI)}
\renewcommand{\basBNFMAP}{Bayesian Neural Field (MAP)}
\renewcommand{\basBNFNS}{Bayesian Neural Field (No Spatial Features)}
\renewcommand{\basSTSVGP}{Sparse Spatiotemporal Variational Gaussian Process}
\renewcommand{\basTSREG}{Trend Surface Regression}
\renewcommand{\basSPDEAR}{Spatiotemporal Generalized Linear Mixed Model (AR1)}
\renewcommand{\basSPDERW}{Spatiotemporal Generalized Linear Mixed Model (RW)}
\renewcommand{\basSPDEIID}{Spatiotemporal Generalized Linear Mixed Model (IID)}
\renewcommand{\basSPDE}{Spatiotemporal Generalized Linear Mixed Model (All)}
\renewcommand{\basGBOOST}{Spatiotemporal Gradient Boosting Trees}
\renewcommand{\basNBEATS}{{Neural Basis Expansion Analysis}}
% Hack because sn-article overwrites table.
\begin{tabular*}{\textwidth}{c} \end{tabular*}
\sisetup{
  propagate-math-font = true,
  text-series-to-math = true,
  round-mode=figures,
  round-precision=2}
\newcommand{\bnum}[1]{\textbf{\num{#1}}}
\newcommand\tstrut{\rule{0pt}{2.4ex}}
\newcommand\bstrut{\rule[-1.0ex]{0pt}{0pt}}
\begin{adjustbox}{max width=1.25\linewidth,center}
\begin{tabular}{|llrrrr|}
\hline\hline
~                & ~               & \multicolumn{3}{c}{\tstrut \textbf{Prediction Error}} & ~ \\ \cline{3-5}
\textbf{Dataset} & \textbf{Method} & \textbf{RMSE}                                 & \textbf{MAE}            & \textbf{MIS}  & \textbf{Runtime}
\tstrut \bstrut \\ \hline \tstrut
Wind Speed
~ & \basBNFVI   & \bnum{2.44}  & \bnum{1.81}       & \bnum{11.88}  & {1167} \\
~ & \basBNFMAP  & \num{2.61}   & \num{1.93}        & \num{12.65}   & {927}  \\
~ & \basSTSVGP  & \num{5.04}   & \num{4.18}        & \num{24.72}   & 1112 \\
~ & \basGBOOST  & \num{3.74}   & \num{2.79}        & \num{18.43}   & 2907 \\
~ & \basNBEATS  & \num{5.20}   & \num{4.07}        & \num{22.92}   & 9237 \\
~ & \basSPDE    & $\xmark$     & $\xmark$          & $\xmark$      & $\xmark$ \\
~ & \basTSREG   & \num{4.94}   & \num{3.88}        & \num{24.83}   & ${\le}1$
\bstrut \\ \hline \tstrut
Air Quality 1
~ & \basBNFVI   & \bnum{5.02}  & \bnum{2.94}       & \bnum{22.52}  & {1169} \\
~ & \basBNFMAP  & \num{5.33}   & \num{3.15}        & \num{24.84}   & {1284} \\
~ & \basSTSVGP  & \num{6.24}   & \num{3.91}        & \num{35.59}   & 1348 \\
~ & \basGBOOST  & \num{7.42}   & \num{4.40}        & \num{31.56}   & 5665 \\
~ & \basNBEATS  & \num{9.23}   & \num{5.95}        & \num{45.11}   & 1461 \\
~ & \basSPDE    & $\xmark$     & $\xmark$          & $\xmark$      & $\xmark$ \\
~ & \basTSREG   & \num{9.35}   & \num{6.62}        & \num{55.98}   & ${\le}1$
\bstrut \\ \hline \tstrut
Air Quality 2
~ & \basBNFVI   & \bnum{8.39}  & \bnum{5.19}       & \bnum{40.08}  & {618} \\
~ & \basBNFMAP  & \num{8.82 }  & \num{5.42}        & \num{43.24}   & {678} \\
~ & \basSTSVGP  & \num{9.92}   & \num{6.78}        & \num{56.12}   & 628 \\
~ & \basGBOOST  & \num{8.77}   & \num{5.57}        & \num{43.71}   & 2671 \\
~ & \basNBEATS  & \num{12.63}  & \num{8.24}        & \num{63.84}   & 778 \\
~ & \basSPDEAR  & \num{11.92}  & \num{7.81}        & \num{73.00}   & 17100 \\
~ & \basSPDERW  & \num{14.62}  & \num{9.48}        & \num{157.10}  & 9447 \\
~ & \basSPDEIID & \num{12.87}  & \num{8.78}        & \num{127.48}  & 3545 \\
~ & \basTSREG   & \num{18.44}  & \num{12.32}       & \num{117.90}  & ${\le}1$
\bstrut \\ \hline \tstrut
Chickenpox Cases
~ & \basBNFVI   & \num{25.96}  & \num{16.09}       & \num{137.74}  & {141} \\
~ & \basBNFMAP  & \num{26.54}  & \num{17.63}       & \bnum{114.44} & {70}  \\
~ & \basSTSVGP  & \num{32.00}  & \num{21.22}       & \num{212.87}  & 63 \\
~ & \basGBOOST  & \num{26.83}  & \num{15.84}       & \num{122.39}  & 189 \\
~ & \basNBEATS  & \num{29.51}  & \num{17.56}       & \num{167.27}  & 250 \\
~ & \basSPDEAR  & \bnum{25.30} & \bnum{15.26}      & \num{179.29}  & 887 \\
~ & \basSPDERW  & \num{26.92}  & \num{16.79}       & \num{179.63}  & 386 \\
~ & \basSPDEIID & \num{28.23}  & \num{16.85}       & \num{327.72}  & 264 \\
~ & \basTSREG   & \num{29.75}  & \num{21.30}       & \num{172.43}  & ${\le}1$
\bstrut \\ \hline \tstrut
Precipitation
~ & \basBNFVI   & \bnum{1.80}  & \bnum{1.23}       & \bnum{8.33}   & {778} \\
~ & \basBNFMAP  & \bnum{1.83}  & \bnum{1.21}       & \bnum{8.28}   & {1069} \\
~ & \basSTSVGP  & \num{3.14}   & \num{2.27}        & \num{31.00}   & {1203} \\
~ & \basGBOOST  & \num{2.63}   & \num{1.67}        & \num{11.13}   & {2064} \\
~ & \basNBEATS  & $\xmark$     & $\xmark$          & $\xmark$      & $\xmark$ \\
~ & \basSPDE    & $\xmark$     & $\xmark$          & $\xmark$      & $\xmark$ \\
~ & \basTSREG   & \num{3.61}   & \num{2.69}        & \num{20.81}   & ${\le}1$
\bstrut \\ \hline \tstrut
Sea Surface Temperature
~ & \basBNFVI   & \num{0.14}   & {0.09}            & \num{0.77}    & {3335} \\
~ & \basBNFMAP  & \bnum{0.10}  & {\textbf{{0.06}}} & \bnum{0.63}   & {4624} \\
~ & \basSTSVGP  & $\xmark$     & $\xmark$          & $\xmark$      & $\xmark$ \\
~ & \basGBOOST  & \num{0.45}   & \num{0.33}        & \num{1.94}    & 12379 \\
~ & \basNBEATS  & \num{0.20}   & \num{0.15}        & \num{0.97}    & 1120 \\
~ & \basSPDE    & $\xmark$     & $\xmark$          & $\xmark$      & $\xmark$ \\
~ & \basTSREG   & \num{0.55}   & \num{0.42}        & \num{2.89}    & 3
\bstrut \\ \hline\hline
\end{tabular}
\end{adjustbox}
\end{table}

%% file: fig-predictions.tex
%!TEX root=./paper.tex
\begin{figure}[p]
\centering
\includegraphics[width=.96\linewidth]{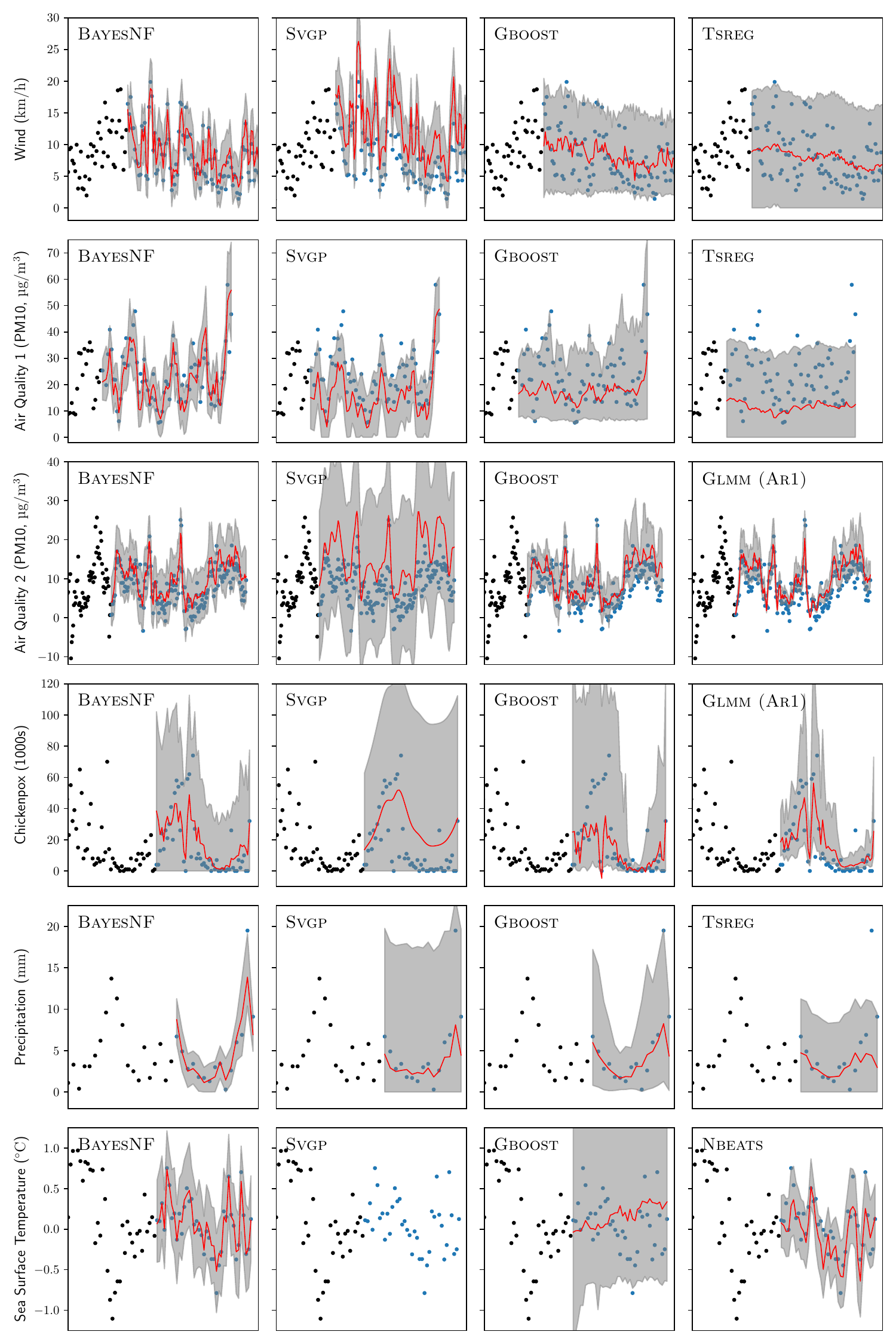}
\captionsetup{skip=0pt}
\caption{Comparison of predictions using \bnf{} and various baselines.
Each row shows results for a given spatiotemporal benchmark dataset
at one spatial location. Black dots are observed data, blue dots are
test data, red lines are median forecasts, and gray regions are
95\% prediction intervals.
(\textsc{BayesNF}: Bayesian Neural Field.
\textsc{Svgp}: Spatiotemporal Sparse Variational Gaussian Process.
\textsc{Gboost}: Spatiotemporal Gradient Boosting Trees.
\textsc{StGLMM}: Spatiotemporal Generalized Linear Mixed Effect Models.
\textsc{NBEATS}: Neural Basis Expansion Analysis.
\textsc{TSReg}: Trend-Surface Regression.)
}
\label{fig:predictions}
\end{figure}

%% file: fig-germany.tex
%!TEX root=./paper.tex
\begin{figure}[p]
\begin{subfigure}{\linewidth}
\includegraphics[width=\linewidth]{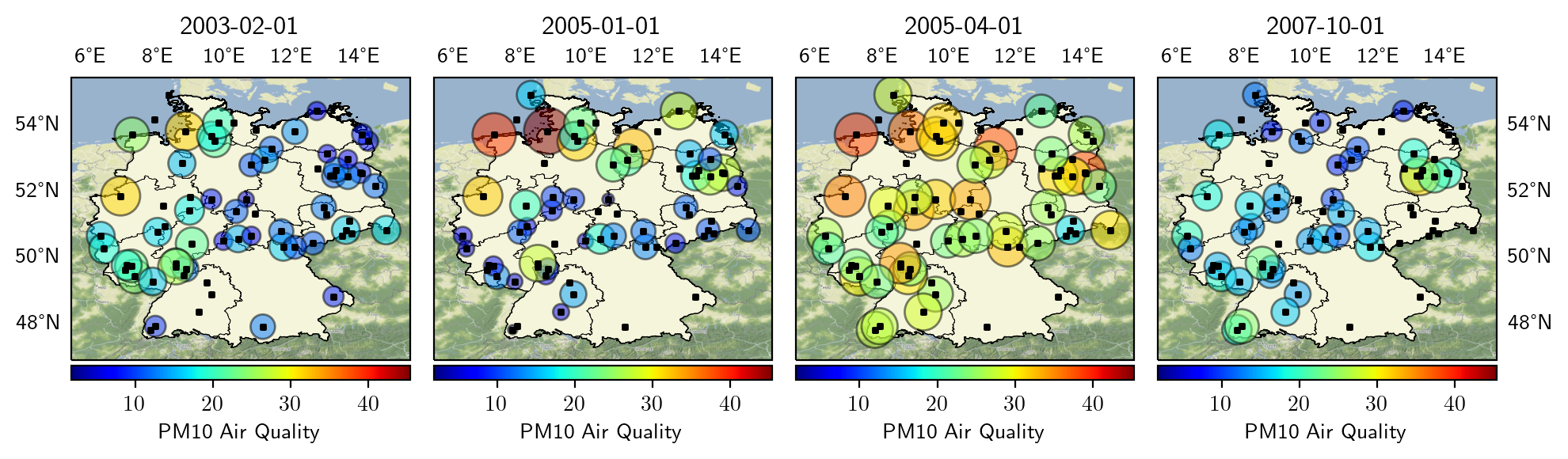}
\captionsetup{belowskip=0pt}
\caption{}
\label{fig:air-data}
\end{subfigure}
\begin{subfigure}{\linewidth}
\includegraphics[width=\linewidth]{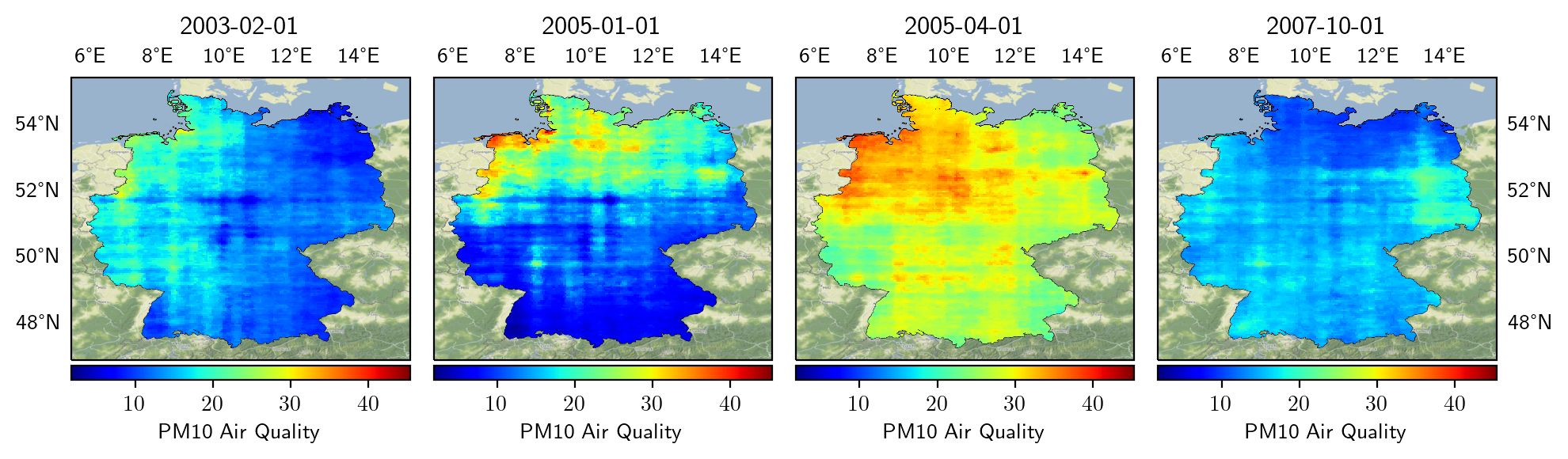}
\captionsetup{belowskip=0pt}
\caption{}
\label{fig:air-spatial}
\end{subfigure}
\begin{subfigure}{\linewidth}
\includegraphics[width=\linewidth]{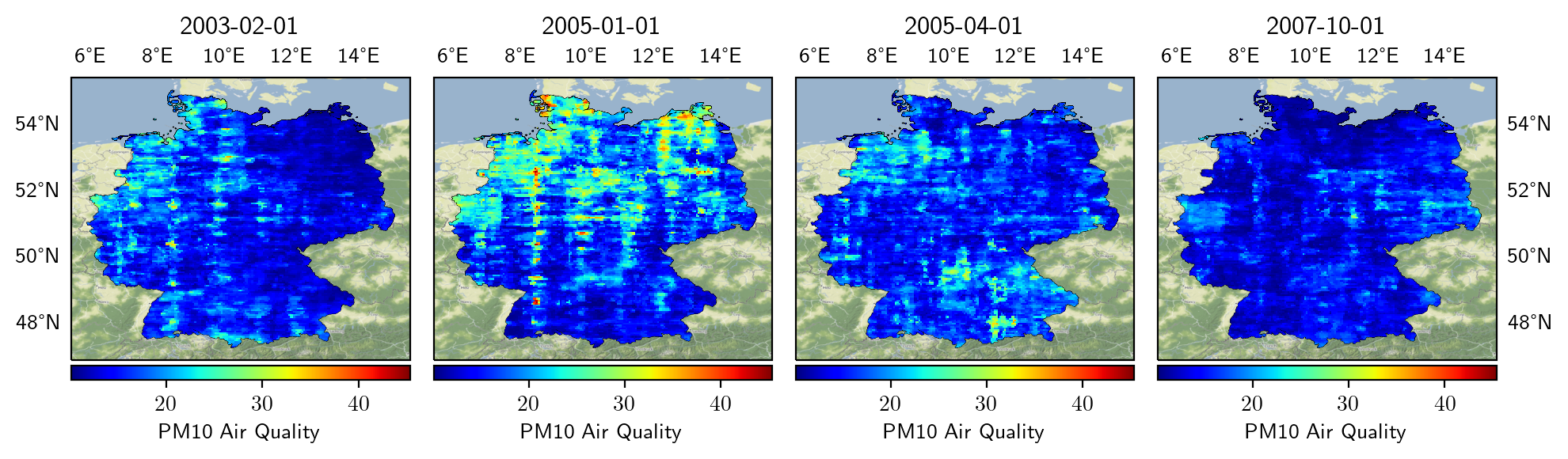}
\captionsetup{belowskip=0pt}
\caption{}
\label{fig:air-spatial-95}
\end{subfigure}
\begin{subfigure}{\linewidth}
\includegraphics[width=\linewidth]{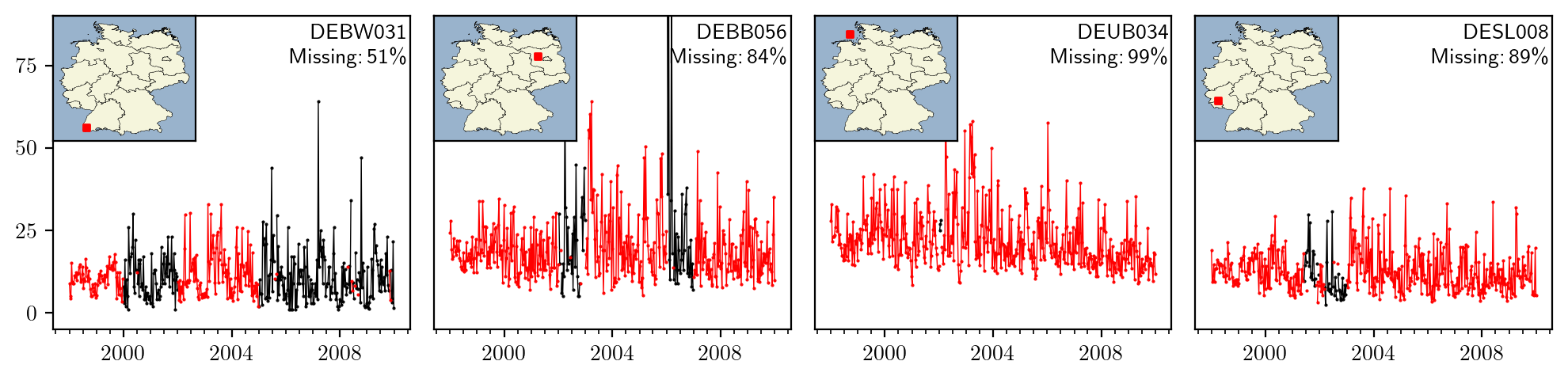}
\caption{}
\label{fig:air-temporal}
\end{subfigure}
\captionsetup{skip=0pt}
\caption{Spatiotemporal prediction of atmospheric particulate matter (PM10) in German air dataset.
\subref{fig:air-data} shows the observed data at four time points:
each shaded circle represents a measurements of PM10 at a given
station. Higher values of PM10 correspond to lower air quality.
The data is sparse: at any given time point, only 47\% of stations (on average) are associated with a PM10 observation.
\subref{fig:air-spatial} Median predictions of PM10 air quality at four time points across the whole spatial field.
\subref{fig:air-spatial-95} Width of 95\% predictions of PM10 air quality at four time points across the whole spatial field.
\subref{fig:air-temporal} Observed PM10 data (black) and median prediction (red) at four sparsely observed locations across time.
Satellite basemap source: \citet{stamen}.
}
\end{figure}

%% file: fig-variogram.tex
%!TEX root=./paper.tex
\begin{figure}[p]
\begin{subfigure}{\linewidth}
\includegraphics[width=\linewidth]{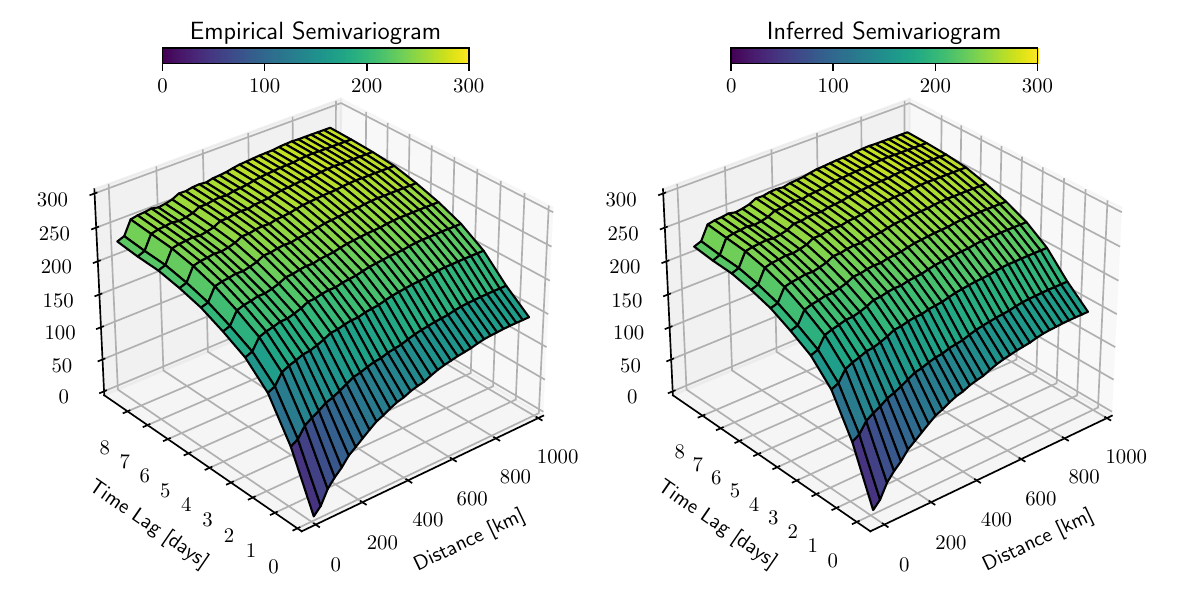}
\caption{}
\label{fig:variogram-top}
\end{subfigure}
\begin{subfigure}{\linewidth}
\includegraphics[width=\linewidth]{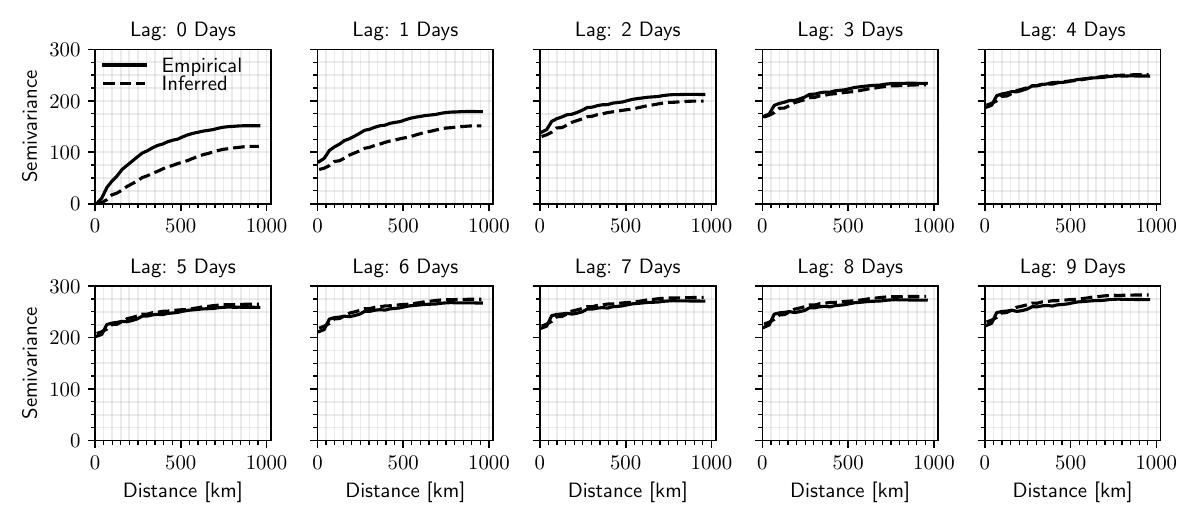}
\caption{}
\label{fig:variogram-bottom}
\end{subfigure}
\caption{Comparison of the empirical and inferred spatiotemporal
semivariograms, which measure the variance of the difference between field
values at a pair of locations, for German PM10 air quality dataset.
The empirical semivariogram is computed using the locations of the 70
stations in the observed dataset.
The inferred semivariogram is computed on 70 novel spatial locations,
sampled uniformly at random within the boundary of the field.
\subref{fig:variogram-top} The agreement between the semivariogram surfaces
indicates that \bnf{} extrapolates the joint spatiotemporal dependence
structure between locations in the observed data to novel locations.
\subref{fig:variogram-bottom} For short time lags less than three days, the
empirical variogram is higher than the inferred variogram at all distances,
showing that \bnf{} models high-frequency day-to-day variance as
unpredictable observation noise.}
\label{fig:variogram}
\end{figure}

%% file: fig-runtime.tex
%!TEX root=./paper.tex
\begin{figure}[p]
\centering
\includegraphics[width=.75\linewidth]{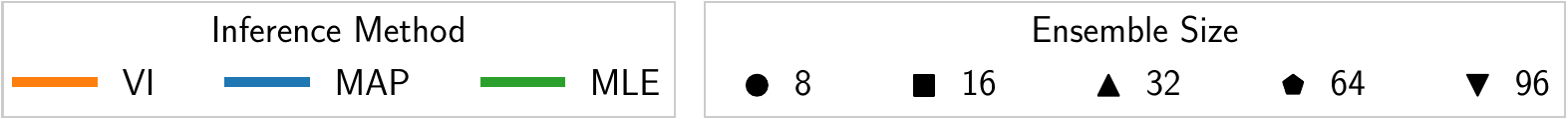}
\begin{tikzpicture}
\node[name=pic]{\includegraphics[width=\textwidth]{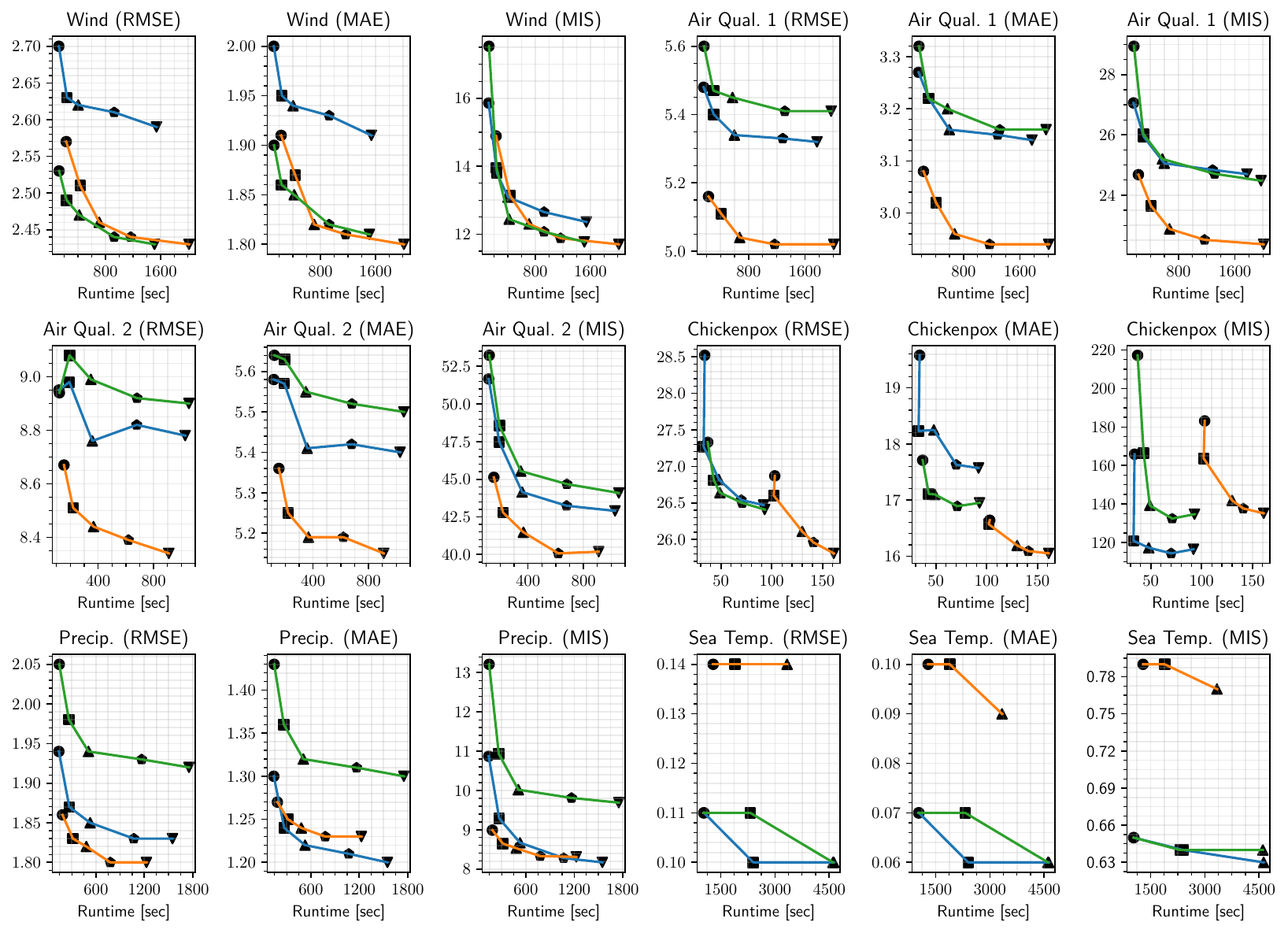}};
\def\offset{50pt}
\draw ($(pic.west) + (0,\offset)$) -- ($(pic.east) + (0,\offset)$);
\draw ($(pic.west) + (0,-\offset)$) -- ($(pic.east) + (0,-\offset)$);
\draw (pic.north) -- (pic.south);
\node[draw=black,rectangle,inner sep=0pt,fit=(pic.north east) (pic.south west)]{};
\end{tikzpicture}
\caption{Runtime versus prediction error profiles using variational inference (VI; orange),
maximum a-posteriori (MAP; blue), and maximum likelihood estimation (MLE;
green) for \bnf{} on the spatiotemporal benchmarks from
\cref{table:datasets}. Markers indicate ensemble size (8, 16, 32, 64, 96).
The predictions errors are given in terms of root-mean square error (RMSE)
and mean average error (MAE) for point forecasts and in terms of mean
interval score (MIS) for 95\% interval forecasts.}
\label{fig:runtime}
\end{figure}

%% file: fig-ablations.tex
%!TEX root=./paper.tex

\newcommand{\abcapt}{%
Percentage change in error (RMSE, MAE, MIS) and runtime
on benchmark datasets using \bnf{} with various modeling ablations.
Horizontal bars in red (resp.~blue) show an increase (resp.~decrease)
in the error and runtime measurements. Errors bars show the minimum
and maximum across all train/test splits.}

\begin{figure}[p]
\centering
\captionsetup[subfigure]{justification=centering,belowskip=10pt}

\begin{subfigure}{\linewidth}
\centering
\includegraphics[width=.85\linewidth]{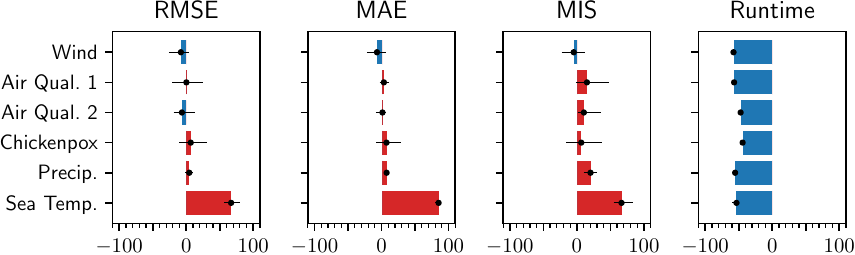}
\caption{Decrease Network Depth By 1}
\label{fig:ablations-depth-down}
\end{subfigure}
\begin{subfigure}{\linewidth}
\centering
\includegraphics[width=.85\linewidth]{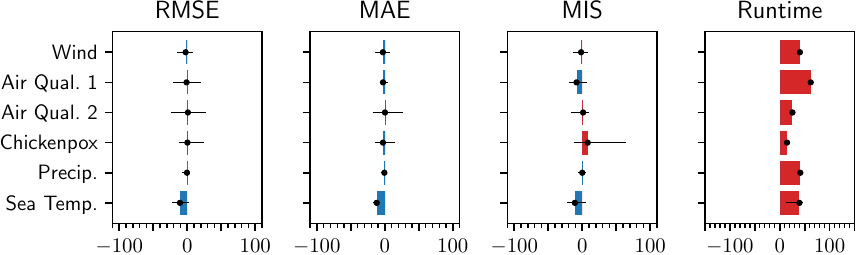}
\caption{Increase Network Depth By 1}
\label{fig:ablations-depth-up}
\end{subfigure}

\begin{subfigure}{\linewidth}
\centering
\includegraphics[width=.85\linewidth]{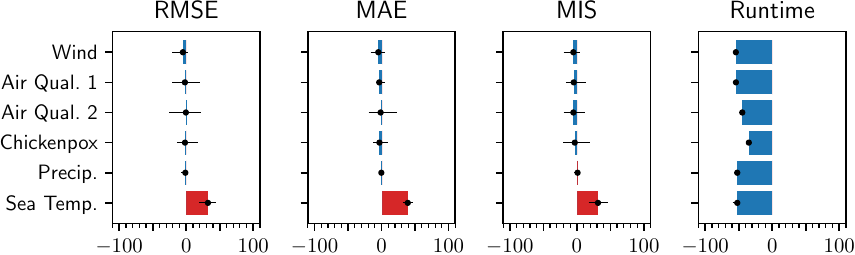}
\caption{Half Network Width}
\label{fig:ablations-width-down}
\end{subfigure}
\begin{subfigure}{\linewidth}
\centering
\includegraphics[width=.85\linewidth]{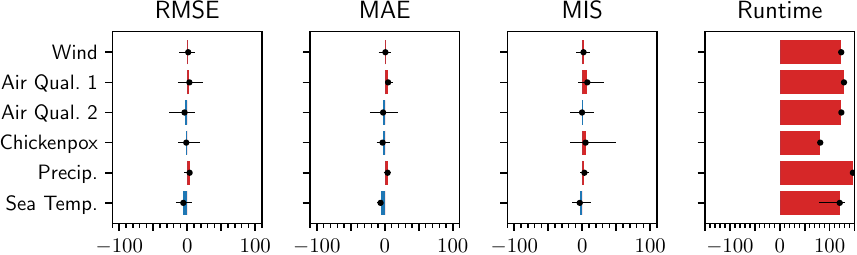}
\caption{Double Network Width}
\label{fig:ablations-width-up}
\end{subfigure}
\captionsetup{skip=0pt}
\caption{\abcapt}
\label{fig:ablations-1}
\end{figure}

\begin{figure}
\ContinuedFloat

\centering
\captionsetup[subfigure]{justification=centering,belowskip=10pt}

\begin{subfigure}{\linewidth}
\centering
\includegraphics[width=.85\linewidth]{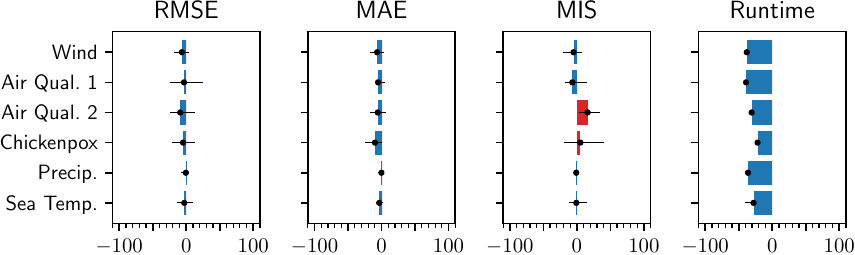}
\caption{No Convex Combination (tanh Activation)}
\label{fig:ablations-tanh}
\end{subfigure}
\begin{subfigure}{\linewidth}
\centering
\includegraphics[width=.85\linewidth]{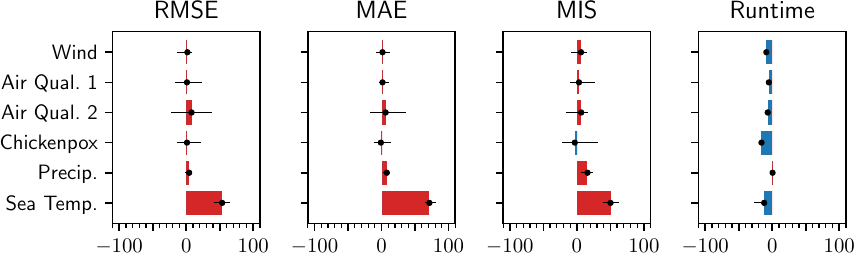}
\caption{No Convex Combination (elu Activation)}
\label{fig:ablations-elu}
\end{subfigure}

\begin{subfigure}{\linewidth}
\centering
\includegraphics[width=.85\linewidth]{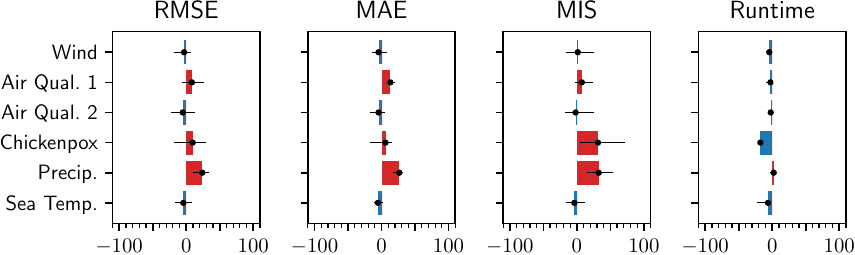}
\caption{No Covariate Scaling Layer}
\label{fig:ablations-scaling}
\end{subfigure}
\begin{subfigure}{\linewidth}
\centering
\includegraphics[width=.85\linewidth]{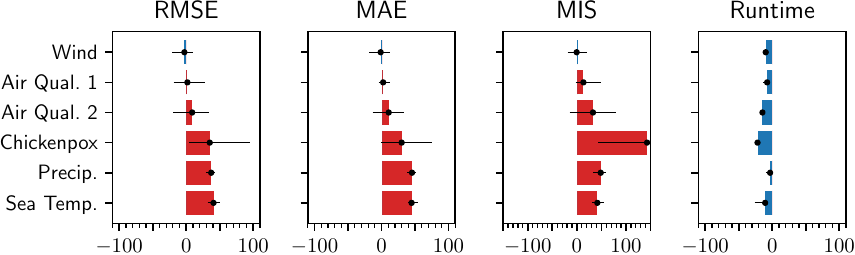}
\caption{No Spatial Fourier Features}
\label{fig:ablations-fourier}
\end{subfigure}

\captionsetup{skip=0pt}
\caption{\abcapt}
\label{fig:ablations-2}
\end{figure}